\pdfoutput=1

\documentclass[11pt]{article}

\usepackage{acl}

\usepackage{times}
\usepackage{latexsym}

\usepackage[T1]{fontenc}

\usepackage[utf8]{inputenc}

\usepackage{microtype}

\usepackage{inconsolata}

\usepackage{graphicx}
\usepackage{booktabs} 
\usepackage{changepage,threeparttable} 
\usepackage{natbib}
\usepackage[utf8]{inputenc} 
\usepackage[T1]{fontenc}    
\usepackage{url}            
\usepackage{booktabs}       
\usepackage{amsfonts}       
\usepackage{nicefrac}       
\usepackage{microtype}      
\usepackage{graphicx}
\usepackage{courier}
\usepackage{algorithm}
\usepackage{algpseudocode}
\usepackage{amsmath}
\usepackage{amssymb}
\usepackage{bbm}
\usepackage{adjustbox}
\usepackage{pifont} 
\usepackage{multirow}
\usepackage{multicol}
\usepackage{changepage,threeparttable} 
\usepackage{makecell}
\usepackage{colortbl}
\usepackage{xspace}
\usepackage{enumitem}
\usepackage{subfigure}
\usepackage{float}
\usepackage{wrapfig}
\usepackage{algorithm}
\usepackage{algorithmicx}

\usepackage{dashrule}
\usepackage{arydshln}

\usepackage{xcolor}
\definecolor{antiquefuchsia}{rgb}{0.57, 0.36, 0.51}
\newcommand{\method}{\textsc{Micro-Act}\xspace}
\definecolor{mycustomcolor}{rgb}{0.68, 0.27, 0.35}
\newcommand{\revise}[1]{\textcolor{black}{#1}}
\title{Micro-Act: Mitigating Knowledge Conflict in LLM-based RAG via Actionable Self-Reasoning}

\author{Nan Huo $^{1}$, Jinyang Li $^{1}$, Bowen Qin $^{2}$\thanks{\quad Corresponding authors are Reynold Cheng and Bowen Qin.} ,
Ge Qu$^{1}$, Xiaolong Li $^{1}$, Xiaodong Li$^{3}$, \\ \textbf{Chenhao Ma} $^{4}$, \textbf{Reynold Cheng} $^{1}$\footnotemark[1]\\
$^{1}$The University of Hong Kong,
$^{2}$ BAAI\\
$^{3}$Xiamen University \\
$^{4}$The Chinese University of Hong Kong, Shenzhen  \\
\texttt{huonan@connect.hku.hk}, \texttt{bwqin@baai.ac.cn}, \texttt{ckcheng@cs.hku.hk} 
}

\begin{document}
\maketitle

\begin{abstract}
Retrieval-Augmented Generation (RAG) systems commonly suffer from \textbf{\textit{Knowledge Conflicts}}, where retrieved external knowledge contradicts the inherent, parametric knowledge of large language models (LLMs). It adversely affects performance on downstream tasks such as question answering (QA). 
Existing approaches often attempt to mitigate conflicts by directly comparing two knowledge sources in a side-by-side manner, but this can overwhelm LLMs with extraneous or lengthy contexts, ultimately hindering their ability to identify and mitigate inconsistencies. To address this issue, we propose \textbf{\method}, a framework with a hierarchical action space that automatically perceives context complexity and adaptively decomposes each knowledge source into a sequence of fine-grained comparisons. These comparisons are represented as actionable steps, enabling reasoning beyond the superficial context.
Through extensive experiments on five benchmark datasets, \method consistently achieves significant increase in QA accuracy over state-of-the-art baselines across all 5 datasets and 3 conflict types, especially in temporal and semantic types where all baselines fail significantly.
More importantly, \method exhibits robust performance on non-conflict questions simultaneously, highlighting its practical value in real-world RAG applications. Code can be found at \url{https://github.com/Nan-Huo/Micro-Act}.
\end{abstract}

\begin{figure}[t]
    \centering
    \includegraphics[width=0.49\textwidth]{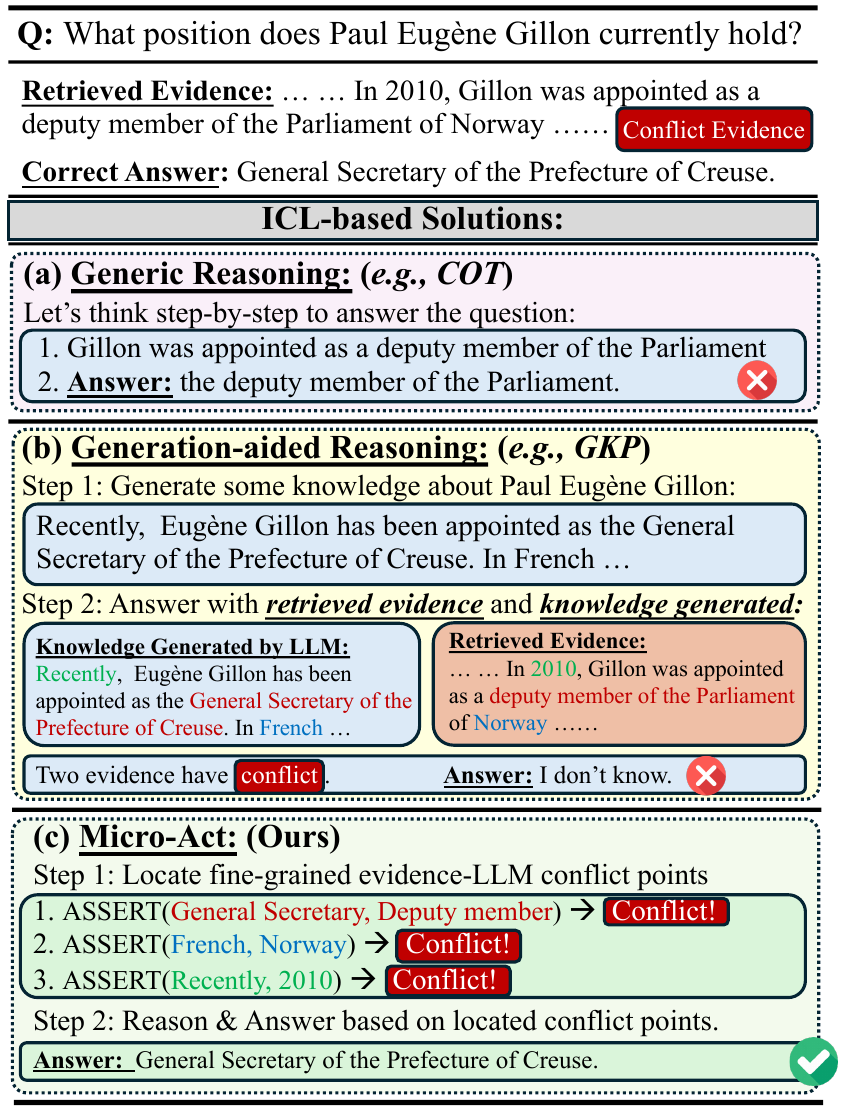}
    \caption{An illustration of QA under knowledge conflict via a real example. The detailed illustration can be found in Figure~\ref{fig:case_study}. (a) refers to the generic reasoning methods that reason on merely retrieved context. (b) refers to generation-aided reasoning methods aided by self-generated knowledge. (c) refers to our proposed~\method.}
    \label{fig:intro}
\end{figure}

\section{Introduction}
Recent advances in large language models (LLMs) have revolutionized natural language processing with their ability to understand and respond to diverse user queries~\citep{chang2024survey,zhao2023survey}. However, relying solely on parametric knowledge often leads to hallucinations and factual errors, especially when dealing with domain-specific queries or rapidly evolving information. To enhance the reliability and factual accuracy of LLM responses, retrieval-augmented generation (RAG) has emerged as a promising paradigm that grounds LLM reasoning with evidence from external knowledge sources~\citep{guu2020retrieval,lewis2020retrieval,chen2024benchmarking, ren2023investigating}. 

Despite the promise of RAG, a critical challenge emerges when retrieved information contradicts the pre-trained parametric knowledge of LLMs, a phenomenon known as \textit{knowledge conflict}~\citep{wang2024astute,jin2024tug}. Such conflicts arise frequently because retrieval systems may introduce noisy, outdated, or even incorrect information~\citep{su2024conflictbank, wang2024astute,shiircan,jin2024tug}, which significantly undermines their potential benefits and raises concerns about their practical deployment in downstream tasks such as question answering (QA).

Prior works addressing knowledge conflicts fall into two distinct categories. The first focuses on specialized fine-tuning techniques~\citep{yuan2024discerning,shiircan,jin2024cutting}. The second leverages In-Context Learning (ICL), which can adapt to new requirements or tasks by providing relevant instructions or examples, reducing the effort required for re-training or continual training. Within the ICL-based category, approaches can be further divided into two types: generic reasoning methods that rely solely on retrieved context, as shown in Figure~\ref{fig:intro}(a), and generation-aided reasoning methods that generate the pre-trained parametric knowledge of LLMs for explicit knowledge comparison with retrieved knowledge~\citep{liu2022generated}, as illustrated in Figure~\ref{fig:intro}(b). However, these ICL-based methods face three critical limitations: (1) heavy reliance on manually crafted instructions limits cross-domain adaptability; (2) side-by-side comparison fails to capture conflicts at different granularity levels, making LLMs vulnerable to irrelevant contexts~\citep{GSMSymbolic}; and (3) those methods meticulously design prompts to handle knowledge conflict, which assumes that knowledge conflict already exists. This would \revise{probably} lead to a negative impact on performance in conflict-free scenarios, which are common in real-world applications, raising concerns about their practical reliability.

To address these limitations, we propose \textbf{\method}, whose core innovation is its ability to dynamically adjust granularity through decomposition action: (1) at model level, it automatically perceives input complexity preferences for different LLMs, and (2) at action level, it detects context granularity of each action and flexibly makes adjustment. As illustrated in Figure~\ref{fig:intro}(c), this adaptive approach enables precise conflict detection across different granularity levels \revise{and reasoning on the conflicts underneath the superficial context.}

Extensive experiments on five widely-used knowledge conflict benchmark datasets grounded in the QA task~\citep{su2024conflictbank,DBLP:conf/iclr/Xie0CL024}, covering diverse knowledge conflict types (mis-information, temporal, and semantic conflicts)~\citep{su2024conflictbank}, demonstrate that \method consistently outperforms state-of-the-art baselines. More importantly, \method also maintains competitive performance in conflict-free cases while state-of-the-art baselines cannot, which underscores the strong robustness of \method. 
Further analysis of complexity detection reveals that \method unlocks the potential of LLMs to perceive complexity and adapt to different environments. \revise{And we find an interesting phenomenon called ``over-rationalization'' which harms conflict resolution and can be mitigated by \method via locating conflict underneath the superficial context.}

These findings validate the effectiveness and robustness of \method in resolving knowledge conflicts for reliable real-world RAG systems.

\section{Related Work}
\paragraph{Retrieval-Augmented Generation for QA.}
Retrieval-augmented generation (RAG) integrates external knowledge sources with language generation, improving the fidelity and robustness of open-domain QA \citep{chen2017drqa,petroni2019lama,asai2020learning,guu2020realm,izacard2020fid,lewis2020retrieval,zhang2024scitat,shi2024exploring}. Subsequent efforts have refined retrieval modules and model architectures to handle diverse knowledge sources and queries more effectively \citep{karpukhin2020dense,izacard2021leveraging,mao2021generation,nakano2021webgpt,shi2023replug,izacard2022atlas,qin2019entity,qin2018end,conforti2020stander,rezaee2024tweetter}. Recent techniques explore dynamic retrieval strategies, domain adaptation, and efficient fine-tuning, further enhancing the adaptability and reliability of RAG frameworks \citep{ram2023incontext,borgeaud2022improving,liu2023lost,zhang2025murre}.

\paragraph{Knowledge Conflict.}
Knowledge conflict surfaces when retrieved evidence disagrees with a model’s internal beliefs or when multiple sources present mutually inconsistent information, resulting in ambiguous or flawed outputs \citep{min2020ambigqa,lewis2020retrieval,shuster2021retrieval,wang2021want,zellers2019defending,tan2024blinded}. This challenge becomes acute in evolving domains (e.g., current events, medicine, science) where timely accuracy is critical \citep{chen2021improving,feng2023factscore}. Conflicts arise not only between retrieved evidence and parametric knowledge but also among multiple retrieved documents, demanding careful reconciliation to avoid misinformation and preserve trustworthiness \citep{thorne2018fact,yang2018hotpotqa,liang2022holistic,vafa2022rarr,shaier2024adaptive,pham2024s,fang2024getting}.

\paragraph{Solutions for Knowledge Conflict.}
Proposed solutions generally follow two broad strategies. The first modifies internal model parameters or architectures to accommodate external evidence more consistently \citep{yuan2024discerning,jin2024cutting,shiircan}, though this often assumes that retrieved information should uniformly override parametric knowledge. The second strategy explicitly identifies and reconciles discrepancies among sources via generation-aided mechanisms or iterative comparison \citep{wang2023resolving,liu2022generated}. While these methods can reduce factual errors, they often rely on ad-hoc instructions or simple pairwise comparisons that fail to capture subtle conflicts. Recent work underscores the need for more principled, robust approaches that integrate nuanced reasoning and validation, improving both fidelity and explainability in retrieval-augmented QA \citep{DBLP:conf/emnlp/XuQGW0ZX24}.

\section{Preliminaries} \label{sec: preliminaries}

\subsection{RAG for QA}
Retrieval-Augmented Generation (RAG) combines a LLM with an external retrieval module. It proceeds in two key phases: a \emph{Retrieval Phase} that returns a set of relevant evidence and a \emph{Generation Phase} where the LLM produces the final answer conditioned on this evidence.

\paragraph{Retrieval Phase.}
Given a query \( q \), a retrieval function \(\mathcal{R}(\cdot)\) returns a set of textual fragments \(\mathcal{E} = \{ e_1, \dots, e_m \}\), where $m$ is the number of fragments. Each fragment \( e_i \) provides potentially relevant information related to \( q \).

\paragraph{Generation Phase.}
Let \(\mathcal{M}_\Theta\) be the LLM parameterized by \(\Theta\). We define the parametric knowledge for the query \(q\) as:
\begin{equation}\label{eq:param_knowledge}
K^{p}(q) = \mathcal{M}_\Theta(q).
\end{equation}
We use \(K^{r}(e_i)\) to represent the knowledge contained in each retrieved fragment \(e_i\). The final answer is produced by conditioning on both the parametric and retrieved knowledge:
\begin{equation}\label{eq:answer_generation}
Ans(q) = \mathcal{M}_\Theta\big(q | K^{p}(q), \{ K^{r}(e_i) \}_{i=1}^m\big).
\end{equation}

\subsection{Knowledge Conflict}
A \emph{knowledge conflict} arises when \(K^{p}(q)\) and some \(K^{r}(e_i)\) are factually or logically inconsistent. Formally, there exists at least one \( e_i \in \mathcal{E} \) such that:
\begin{equation}\label{eq:conflict_condition}
K^{p}(q) \;\not\approx\; K^{r}(e_i),
\end{equation}
where \(\not\approx\) denotes a factual or logical inconsistency. 

\begin{figure*}[t]
    \centering
    \includegraphics[width=1.0\textwidth]{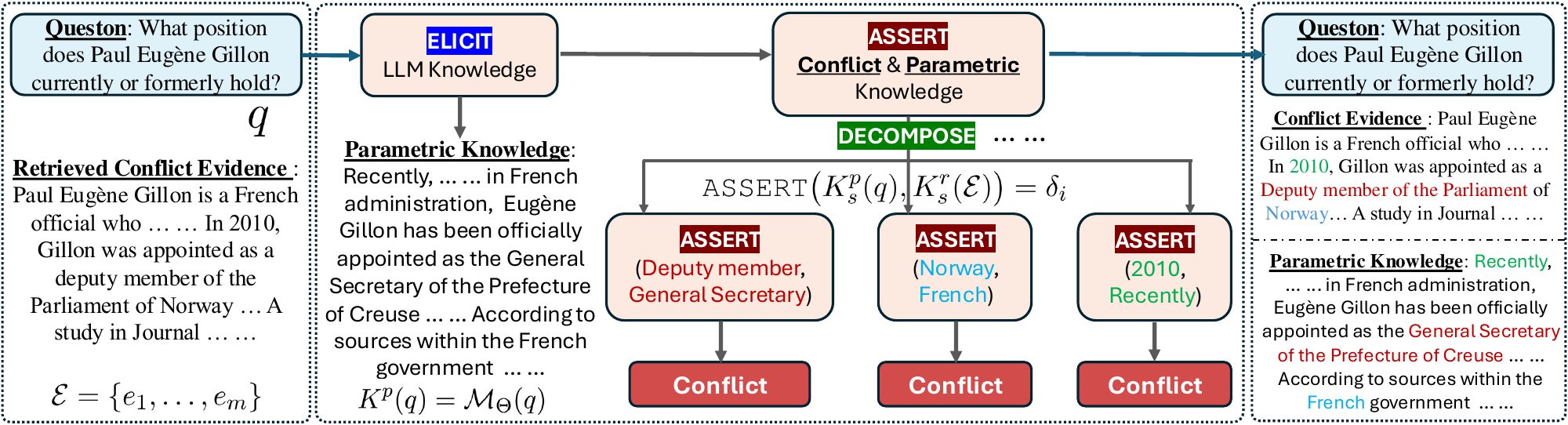}
    \caption{An illustration of handling knowledge conflict in QA task. Actions highlighted with blue color represent navigational actions; Red color represents functional actions; and green color represents the bridging action. "... ..." represents multiple interplayed actions are folded for simplicity.}
    \label{fig:method}
\end{figure*}

\begin{algorithm}[h!]
\caption{\method\ Pseudocode}
\begin{algorithmic}[1]
\State \textbf{Input:} query $q$, external corpus $\mathcal{E}$, LLM $\mathcal{M}_\Theta$, turn budget $N$
\State \textbf{Retrieve:} $K^{P}\!\gets\!{\tt ELICIT}(q)$, \;$K^{r}\!\gets\!{\tt RETRIEVE}(\mathcal{E},q)$
\State $H\!\gets\!\emptyset$
\For{$t=1$ \textbf{to} $N$}
    \State $T_t\!\gets\!\mathcal{M}_\Theta(\cdot\mid H)$; \; $A_t\!\gets\!{\tt SELECT}(T_t)$
    \State $O_t\!\gets\!
        \begin{cases}
          {\tt REASON}(\cdot)\\
          {\tt ASSERT}(\cdot)\\
          {\tt DECOMPOSE}(\cdot)
        \end{cases}$
    \State $H\!\gets\!H\cup\{T_t,A_t,O_t\}$
    \If{$O_t=\text{conflict}\ \land\ {\tt COMPLEX}$}
        \State $A_t\!\gets\!\textsc{Decompose}$   \Comment{force split}
    \EndIf
    \If{{\tt SOLVED}$(H)$} \textbf{break}
    \EndIf
\EndFor
\State \textbf{Return} $\mathcal{M}_\Theta(\textsc{Answer}\mid H)$
\end{algorithmic}
\label{algo}
\end{algorithm}

\section{Methodology} \label{sec: methodology}
We introduce \textbf{\method}, a framework that enables Large Language Models (LLMs) to automatically identify and resolve detailed points of knowledge conflict. \method comprises three key components: (1) a hierarchical action space (Section \ref{sec:act_sapce}), (2) a Reasoning Body (Section \ref{sec:reasoning_body}), and (3) Adaptive Granularity and Optimization strategies (Section \ref{sec:adapt_granularity} and \ref{sec:infinite-decomp}). The detailed pseudocode of \method can be found in Algorithm~\ref{algo}.

\subsection{Hierarchical Action Space} 
\label{sec:act_sapce}

Establishing a well-structured action space allows LLMs to more efficiently invoke planning strategies \citep{yao2024tree}. To this end, we define the action space as a structured integration of three key categories: (1) \textit{navigational actions}, (2) \textit{functional actions}, and (3) \textit{bridging actions}, with the decomposition component serving as the cornerstone for refining context granularity of actions.

\paragraph{Navigational Actions.} They focus on exploring the environment and obtaining more information as the prerequisite of effective reasoning~\citep{gu2024middleware}. Navigational actions include eliciting parametric knowledge from the LLM and getting the reasoning path of a QA task based on input context. Let \(\mathcal{A}_{\text{nav}}\) represent \textit{navigational actions}. Specifically, we formally define the elicit action in Eq.~\ref{eq:elicit_action}.
\begin{equation}\label{eq:elicit_action}
\texttt{ELICIT}(q) = K^{p}(q) = \mathcal{M}_\Theta(q).
\end{equation}

And we formally define the action to get the reasoning path $\mathcal{P}_K$ in Eq.~\ref{eq:reason}.
\begin{equation}\label{eq:reason}
\texttt{REASON}(K) = \mathcal{P}_K = \mathcal{M}^p_\Theta(K),
\end{equation}
where $\mathcal{M}^p_\Theta(K)$ represents prompting LLM parametrized by $\Theta$ to generate a reasoning path on $K$. And $K$ is the input knowledge representation either from $K^{p}(q)$ or from $K^{r}(\mathcal{E})$. 

\paragraph{Functional Actions.} They address conflict detection either between retrieved evidence and LLM parametric knowledge or between their reasoning paths generated by the navigational action. Once relevant information is prepared, \textit{functional actions}, denoted by \(\mathcal{A}_{\text{func}}\), detect conflict among them. Formally, we define the assert action to implement this logic, which is a conflict verification action and checks the consistency between \(K^{p}(q)\) and a particular \(K^{r}(\mathcal{E})\) in Eq.~\ref{eq:func_conflict}.
\begin{equation}\label{eq:func_conflict}
\texttt{ASSERT}\bigl(K_s^{p}(q), K_s^{r}(\mathcal{E})\bigr) = \delta_i,
\end{equation}
where \(\delta_i \in \{0, 1\}\). If \(\delta_i = 1\), a conflict is detected. And $K_s^{p}(q) \in K^{p}(q)$ means $K_s^{p}(q)$ is a partial knowledge of $K^{p}(q)$. 

\paragraph{Bridging-Action.} It is responsible for dynamically optimizing granularity by decomposing actions when needed. A side-by-side assert action may fail to detect subtle conflicts embedded in lengthy, noisy contexts. To address this, we introduce the \textit{decomposition action}, collected in \(\mathcal{A}_{\text{micro}}\), which can refine the granularity of analysis. Suppose an $\texttt{ASSERT}(\cdot)$ action on complex knowledge context is represented as $\texttt{ASSERT}\bigl(K^{p}(q), K^{r}(\mathcal{E})\bigr)$. A decomposition action can decompose this complex reasoning into smaller, manageable action steps, as shown in Eq.~\ref{eq:meta_decompose}.
\begin{equation}\label{eq:meta_decompose}
\begin{aligned}
& \texttt{DECOMPOSE}(\texttt{ASSERT}\bigl(K_s^{p}(q), K_s^{r}(\mathcal{E}\bigr)) \\
& = \{\texttt{ASSERT}\bigl(K_{s'}^{p}(q), K_{s'}^{r}(\mathcal{E})\bigr), \dots \},
\end{aligned}
\end{equation}
where $K_s^{p}(q) \in K^{p}(q)$ is a partial knowledge of $K^{p}(q)$. And $K_{s'}^{p}(q)$ refers to $K_s^{p}(K_s^{p}(q))$, which means the finer-grained partial knowledge of $K_s^{p}(q)$.
Each newly created sub-action deals with a further fragment of the evidence, increasing the likelihood of revealing fine-grained conflicts. It will decompose the action until LLM has enough confidence or reach the max turn limit.

\subsection{Reasoning Body}
\label{sec:reasoning_body}

We integrate our hierarchical action space with the ReAct process~\citep{react_yao} to teach LLM integrate our hierarchical action space to automatically handle knowledge conflicts. At step \( t \), the LLM first produces a thought \( T_t \):
\begin{equation}
T_t \sim \mathcal{M}_\Theta(T_t \mid H_{t-1}),
\end{equation}
where \( H_{t-1} \) is the accumulated history of all thoughts, actions, and observations before step \( t \).

Conditioned on \( H_{t-1} \) and the newly generated thought \( T_t \), the model selects an action \( A_t \):
\begin{equation}
A_t \sim \mathcal{M}_\Theta(A_t \mid H_{t-1}, T_t).
\end{equation}
This action, executed in the changing environment (for example, the knowledge has been decomposed at different granularity), yields an observation \( O_t \):
$O_t = \text{Env}(A_t).$ 
The history is then updated:
\begin{equation}
H_t = H_{t-1} \cup \{T_t, A_t, O_t\}.
\end{equation}

This iterative process continues, adjusting granularity via decomposition actions whenever subtle conflicts require finer checks. After \( N \) steps, the final answer is generated:
\begin{equation}
A_f \sim \mathcal{M}_\Theta(A_f \mid H_N).
\end{equation}

By dynamically selecting navigational, functional, and decomposition actions, this procedure ensures subtle knowledge conflicts are detected and mitigated, improving the reliability of the final output. An example illustration of this process is shown in Figure~\ref{fig:method}.

\section{Understanding Complexity-Driven Knowledge Decomposition Dynamics}
\label{sec:adapt_granularity}
To gain a deeper understanding of how model bridging actions are related to complexity, we follow \citep{murtybagel} to characterize the distribution of the newly inferred knowledge representation at turn $t$ based on trajectory over previous $t-1$ turns. Specifically, we define:
\begin{equation}
\begin{aligned}
    &p_t(K_n) = \\
    &\sum_{c'} \sum_{K} p_{\text{model}}(K_n \mid c')\, p_{\text{verify}}(c' \mid K)\, p_{t-1}(K),
\end{aligned}
\label{p_model}
\end{equation}
where \(K\) is current knowledge representation, \(K_n\) is the newly inferred knowledge representation (often obtained by decomposing \(K\)), \(c\) is the ground-truth knowledge conflict, and \(c'\) is a potentially incorrect knowledge conflict identified by the model. $p_{\text{model}}$ means the distribution on generate new knowledge. $p_{\text{verify}}$ means the distribution on generating conflicts. Detailed derivation can be found in Appendix~\ref{sec:appendix_derivation}.

In this formulation, the term \(\sum_{K} p_{\text{verify}}(c' \mid K)\) increases with the complexity (e.g., longer context, harder domain and etc.), resulting in higher verification probabilities and an increased risk of inaccurate conflict detection. And \(\sum_{c'} p_{\text{model}}(K_n \mid c')\) depends on the LLM compatibility. A less capable LLM is more likely to be influenced by erroneous conflicts (\(c'\)), thereby requiring further decomposition and pushing \(p_k(K_n)\) higher. Section~\ref{sec:exp_complexity_analysis} shows more details about how these factors drive proactive decomposition across models.

\section{Preventing Infinite Decomposition}
\label{sec:infinite-decomp}

While hierarchical reasoning is essential for resolving complex conflicts, an unconstrained recursive process could, in principle, keep splitting a context. Building upon the probabilistic dynamics in Eq.~\eqref{p_model}, we show that \method\ can prevent infinite decomposition, and we complement this theoretical safeguard with a hard maximum turn budget.

\paragraph{Complexity‐Aware Stopping Criterion.}
Let \(\mathcal{C}_t\) denote the latent \emph{complexity score} of the current context after \(t\) turns.  A decomposition step is triggered only when \(\mathcal{C}_t > \tau\), where \(\tau\) represents the minimum complexity the underlying LLM can handle confidently.  
Because each decomposition shortens the context length and narrows its semantic scope, the following strict inequality holds:
\begin{equation}
\mathcal{C}_{t+1} < \mathcal{C}_t, 
\qquad \forall\, t\ge0.
\label{eq:complexity_monotone}
\end{equation}
Define 
\begin{equation}
T_\tau \;=\; 
\min\bigl\{\,t \mid \mathcal{C}_t \le \tau \bigr\}.
\end{equation}
By Eq.~\eqref{eq:complexity_monotone}, \(T_\tau\) is finite, and once reached we have \(p_t(K_n)=0\); no further actions in the \textsc{Decompose} branch will be sampled. In other words, the process is \emph{self‐regularising}: an LLM that already “understands’’ the context (small \(\mathcal{C}_t\)) simply refuses to split it further.

\begin{table*}[t]
\centering
\footnotesize
\small
\resizebox{\linewidth}{!}{
\begin{tabular}{lcccccccc}
\toprule
\multirow{2}[4]{*}{\textbf{Prompting}} & \multicolumn{2}{c}{\textbf{GPT-4o}} & \multicolumn{2}{c}{\textbf{GPT-4o-mini}} &
\multicolumn{2}{c}{\textbf{LLaMA-3.1-70B}} &
\multicolumn{2}{c}{\textbf{LLaMA-3.1-8B}}
\\
\cmidrule(lr){2-3} \cmidrule(lr){4-5} \cmidrule(lr){6-7}  \cmidrule(lr){8-9} &
\textbf{ConflictBank} & \textbf{KRE} & \textbf{ConflictBank} & \textbf{KRE} & \textbf{ConflictBank} & \textbf{KRE} & \textbf{ConflictBank} & \textbf{KRE} \\
\midrule
\textbf{\textit{Generic Reasoning}} \\
\quad End-to-End QA &   5.40   &  43.80 & 2.77 & 31.10 & 3.07 & 14.50 & 2.53 & 9.55  \\
\quad Few-Shot QA & 6.30  & 45.65 & 2.83 & 33.30 & 3.87 & 15.20 & 3.13 & 10.30  \\
\quad Chain-of-Thought~\citep{wei2022chain} &  6.43  &  44.35 & 3.00 & 36.50 & 1.40 & 29.45 & 2.13 & 24.50 \\
\midrule
\textbf{\textit{Generation-aided Reasoning}} \\
\quad Self-Ask~\citep{press2023measuring} & 3.13 & 41.45& 2.57 & 24.90 & 3.33 & 23.65 & 2.77 & 18.65 \\
\quad Comparative~\citep{wang2023resolving} & 11.70 & 33.95 & 2.10 & 23.85 & 4.53 & 25.25 & 3.87 & 19.80 \\
\quad GKP~\citep{liu2022generated} &  \underline{15.40}  &  \underline{55.30} & \underline{17.53}  & \underline{44.45} & \underline{15.83} & \underline{43.55} & \underline{6.83} & \underline{32.75} \\
\midrule
\quad~\method (ours) & 
\textbf{22.30 \scriptsize{\textcolor{antiquefuchsia}{($\uparrow$ 6.90)}}} & 
\textbf{59.50 \scriptsize{\textcolor{antiquefuchsia}{($\uparrow$ 4.20)}}} & 
\textbf{26.93 \scriptsize{\textcolor{antiquefuchsia}{($\uparrow$ 9.40)}}} & 
\textbf{51.10 \scriptsize{\textcolor{antiquefuchsia}{($\uparrow$ 6.65)}}} & 
\textbf{26.50 \scriptsize{\textcolor{antiquefuchsia}{($\uparrow$ 10.67)}}} & 
\textbf{54.90 \scriptsize{\textcolor{antiquefuchsia}{($\uparrow$ 11.35)}}} & 
\textbf{18.30 \scriptsize{\textcolor{antiquefuchsia}{($\uparrow$ 11.47)}}} & 
\textbf{46.60 \scriptsize{\textcolor{antiquefuchsia}{($\uparrow$ 13.85)}}} \\
\bottomrule
\end{tabular}
}
\caption{The average results of Question Answering under Knowledge Conflict on ConflictBank and KRE with GPT-4o-mini, GPT-4o, LLaMA-3.1-70B and LLaMA-3.1-8B. The performance is on average over its sub-datasets. (\underline{underline} denotes the previous SOTA performance; \textbf{bold} denotes the best performance; the improvement \textcolor{antiquefuchsia}{($\uparrow$)} is measured against the previous SOTA performing method.)}
\label{tab:overall_results}%
\end{table*}%

\section{Experiments}
\subsection{Experiment Settings}
\paragraph{Datasets.}
We evaluate \method on five benchmark datasets drawn from two comprehensive collections: ConflictBank and KRE. ConflictBank~\citep{su2024conflictbank} provides three specialized datasets targeting distinct conflict types: \textbf{misinformation}, \textbf{temporal} discrepancies, and \textbf{semantic} divergences between retrieved and parametric knowledge. From KRE~\citep{ying2023intuitive}, we utilize {MuSiQue\_KRE} and {SQuAD\_KRE}, derived from MuSiQue~\citep{DBLP:journals/tacl/TrivediBKS22} and SQuAD v2.0~\citep{DBLP:conf/acl/RajpurkarJL18} respectively. These datasets feature multiple-choice questions with generated explanations supporting incorrect choices, creating controlled scenarios for examining reasoning conflicts. Due to the limitation of computational resources, we randomly sampled 3000 data in ConflictBank and 2000 data in KRE dataset across all features, and corrected any errors found. 

\paragraph{Metrics \& Models.}
Following existing knowledge conflict works~\citep{DBLP:conf/iclr/Xie0CL024, su2024conflictbank,wang2023resolving, shiircan}, we measure knowledge conflict in QA task by employing QA accuracy as our primary evaluation metric. Specifically, the answer format of QA is multiple-choice. If LLMs successfully resolve knowledge conflict, they will choose the correct answer instead of the negative answer supported by the conflict (wrong) knowledge~\citep{su2024conflictbank}.

In our experiments, we use GPT-4o, GPT-4o-mini~\citep{openai2023gpt4}, LLaMA-3.1-70B and LLaMA-3.1-8B~\citep{Dubey2024TheL3} as the backbone LLMs.

\paragraph{Compared Methods.}
We evaluate \method against two categories of ICL-based approaches: generic reasoning methods that reason on retrieved evidence, including end-to-end QA, few-shot QA, and COT~\citep{wei2022chain}; and generation-aided reasoning methods that reason with self-generated content of LLMs, including Self-Ask~\citep{press2023measuring}, GKP~\citep{liu2022generated}, and Comparative~\citep{wang2023resolving}. We evaluate these methods across all five datasets from ConflictBank and KRE. Prompts and implementation details can be found in Appendix~\ref{imp_detail}.

\paragraph{Implementation.}
We implement~\method using zero-shot prompting without task-specific customization. To ensure reproducibility, we maintain consistent parameters across all experiments: temperature = 0, top-p = 1, and maximum generation length = 512 tokens (\texttt{max\_tokens} for closed-source LLMs, \texttt{max\_new\_tokens} for open-source models). We utilize the Hugging Face Transformers library for open-source model inference. All experiments with open-source models are conducted on 4 NVIDIA A100 GPUs (80GB), while closed-source models are accessed via their respective API endpoints.

\begin{figure}[h]
    \centering
    \includegraphics[width=0.48\textwidth]{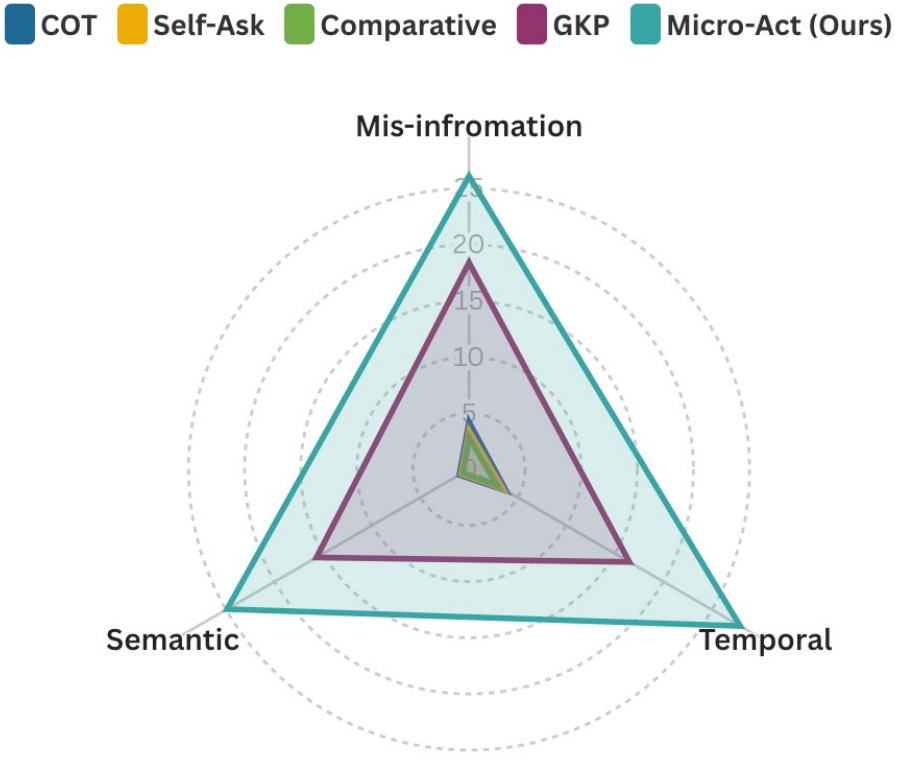}
    \caption{The detailed performance of~\method across all 3 conflict types with GPT-4o-mini.}
    \label{fig:radar}
\end{figure}

\subsection{Main Results} \label{sec: main_res}
We summarize the performance of \textbf{\method} and various baseline methods on ConflictBank and KRE in Table~\ref{tab:overall_results}. And detailed performance comparison across all three conflict types (i.e., mis-information, temporal, and semantic) is shown in Figure~\ref{fig:radar}.

\method surpasses all baseline approaches across all tested LLMs. Notably, \method improves over the previous SOTA method by up to 9.40\% on ConflictBank and 6.65\% on KRE for GPT-4o-mini, and by 11.47\% and 13.85\% on LLaMA-3.1-8B, respectively. Results across all 5 datasets and 3 conflict types, confirm the superior capability of \method in handling knowledge conflict and suggest that such superior capability is not model-specific.

\subsection{Over-Rationalization Issue} \label{sec: over}
In our experiments, we observed an intriguing phenomenon: when presented with both conflicting evidence and LLMs parametric knowledge, LLMs sometimes attempt to \textit{support all contradictory information as equally valid}. We characterize this behavior as ``\textbf{over-rationalization}'', which is a tendency to find complex justifications that make contradictory evidence appear compatible. Surprisingly, more capable models like GPT-4o exhibit this behavior more frequently than GPT-4o-mini, leading to performance degradation in GKP as shown in Table~\ref{tab:overall_results}. 

\revise{Furthermore, we observe that the issue of ``over-rationalization'' is strongly associated with the type of conflict, occurring more frequently in temporal and semantic conflicts. Unlike misinformation-based conflicts, where conflicts are typically explicit and directly presented in the context, the temporal and semantic conflicts are often implicit beneath the superficial context, misleading LLMs to rationalize both sides of conflict. A detailed case analysis is in Section~\ref{sec: case_study}.}

\revise{However, \method can ``visualize'' the underlying reasoning path via dynamic decomposition to pinpoint finer-grained conflict and focus on those \textit{nuanced conflicts underneath the superficial meaning of context}. Those conflicts cannot be effectively detected through simple side-by-side comparisons used by baseline methods. As illustrated in Figure~\ref{fig:radar}, \method achieves a more significant performance improvement over baselines specifically in the \textit{Temporal} and \textit{Semantic} conflict types. Detailed analysis is in Appendix~\ref{app: Over-Rationalization}.}

\begin{figure}[h]
    \centering
    \includegraphics[width=0.48\textwidth]{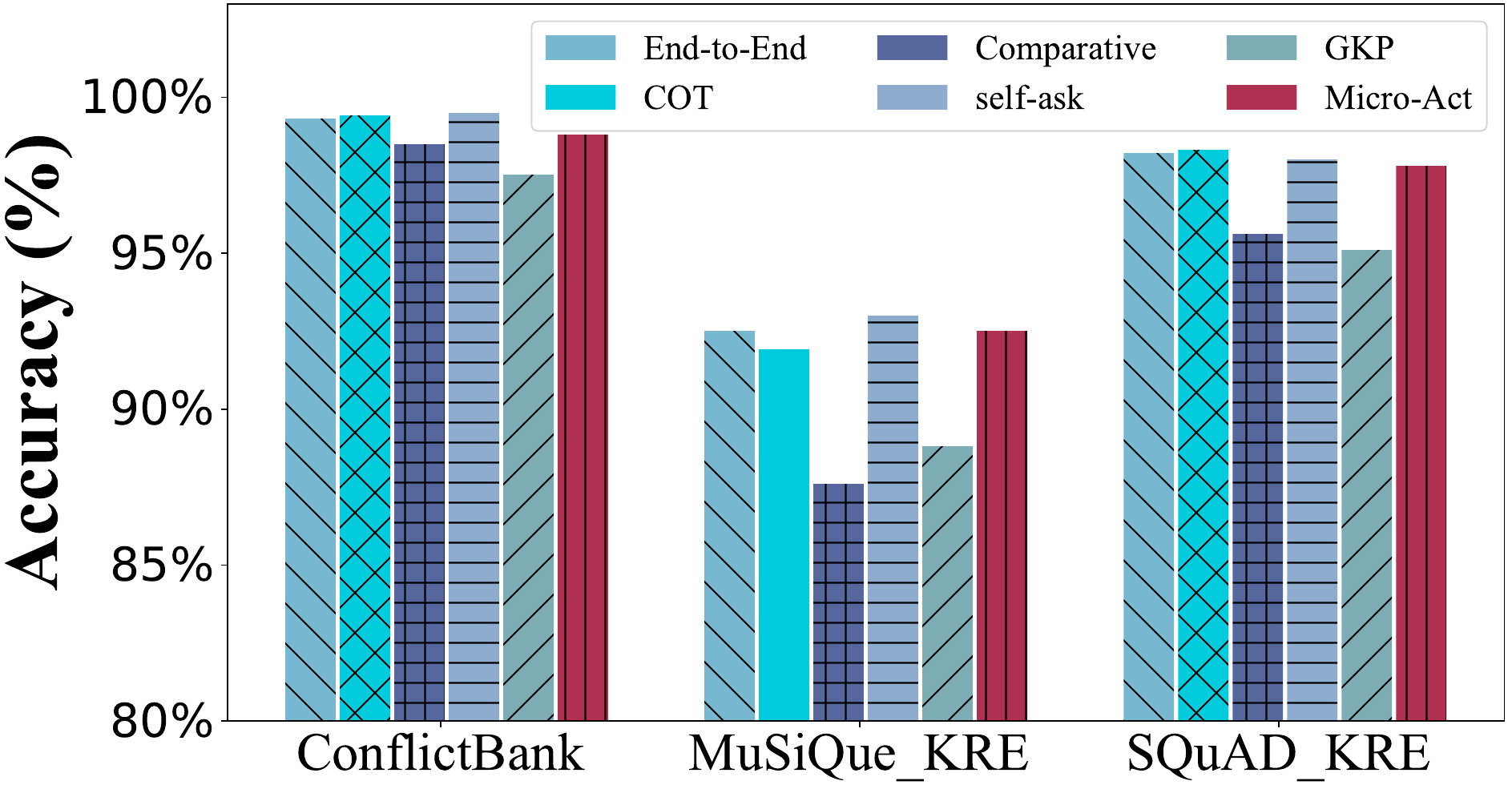}
    \caption{The performance of~\method and baselines using GPT-4o-mini under QA task \textbf{without knowledge conflict}.}
    \label{fig:robustness}
\end{figure}

\begin{figure*}[t]
	
	\begin{minipage}{0.32\linewidth}
		\vspace{3pt}
		\centerline{\includegraphics[width=\textwidth]{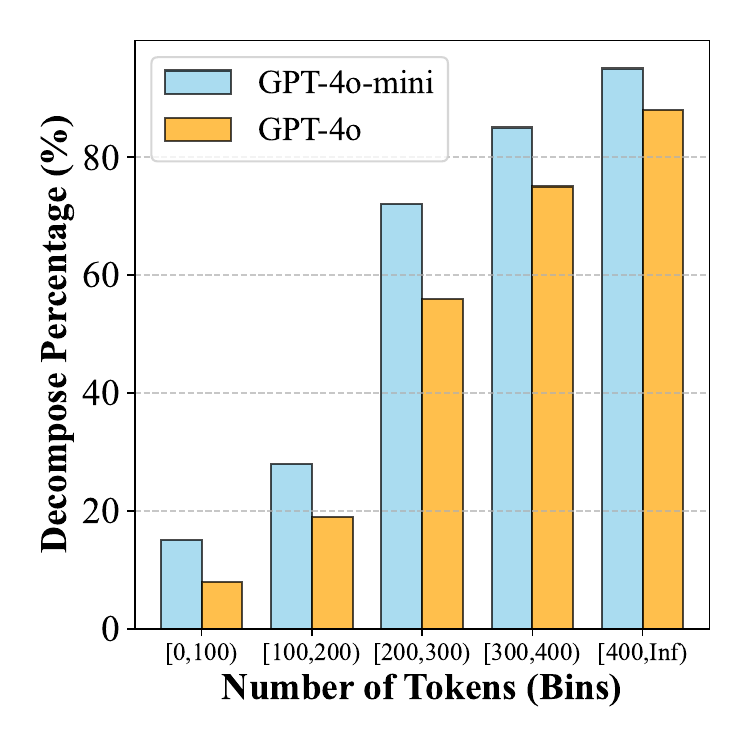}}
		\centerline{(a) Different Context Length}
	\end{minipage}
	\begin{minipage}{0.32\linewidth}
		\vspace{3pt}
		\centerline{\includegraphics[width=\textwidth]{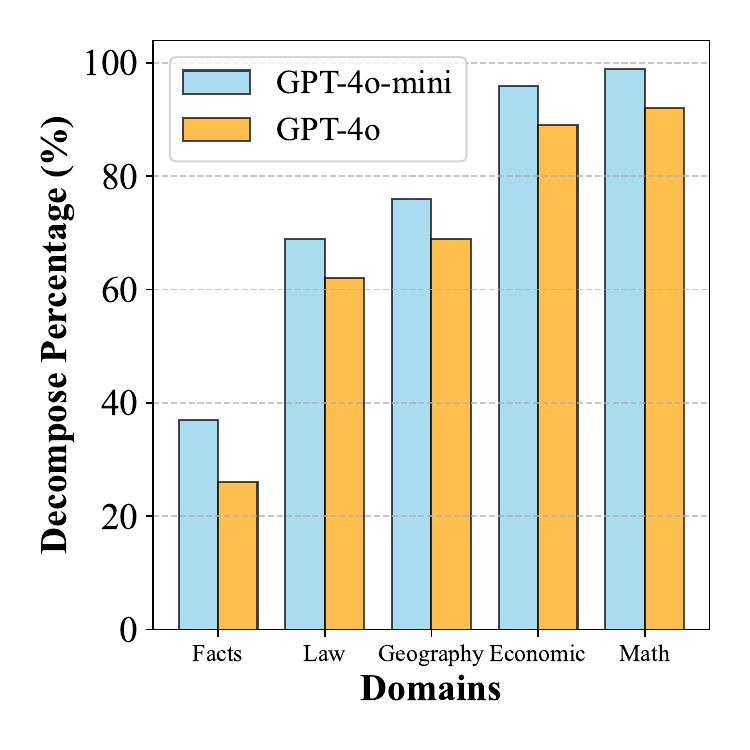}}
	 
		\centerline{(b) Different Question Domains}
	\end{minipage}
	\begin{minipage}{0.32\linewidth}
		\vspace{3pt}
		\centerline{\includegraphics[width=\textwidth]{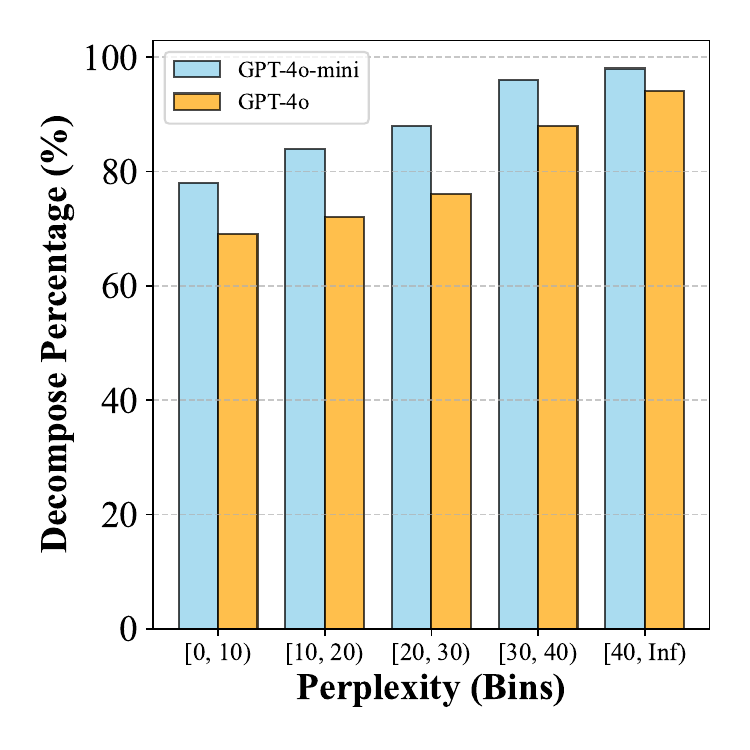}}
	 
		\centerline{(c) Different Context Perplexity}
	\end{minipage}
 
	\caption{The visual comparisons of the \texttt{DECOMPOSE} action utilization percentage in different complexity, including different context length, question domains and perplexity using GPT-4o-mini and GPT-4o. The detailed calculation of perplexity can be found in Appendix~\ref{sec:perp_calc}.}
	\label{fig:complexity_analysis}
\end{figure*}

\subsection{Robustness Under Conflict-Free Scenarios}
\label{sec:exp_robustness_no_conflict}

Many conflict resolution methods assume the presence of knowledge conflicts. However, in real-world applications, it is often impossible to pre-determine whether retrieved content conflicts with the parametric knowledge of LLMs, making robustness in conflict-free scenarios crucial.

\revise{As shown in Figure~\ref{fig:robustness}, existing approaches face a trade-off. Generic reasoning methods like end-to-end and COT achieve high accuracy in conflict-free cases but degrade significantly (by 70-95\%) when conflicts arise. And generation-aided methods such as GKP improve conflict resolution but exhibit lower accuracy in conflict-free cases.}

\method overcomes this limitation by achieving state-of-the-art performance in conflict scenarios with over 24\% performance gain and showing robustness with only sacrificing less than 2\% accuracy in conflict-free cases, compared with the end-to-end or self-ask baseline. Rather than introducing biases to favor certain evidence sources, \method helps models automatically identify and analyze potential conflicts through structured action space with decomposition, enabling robust performance regardless of whether conflicts exist.

\subsection{Complexity Perception Analysis}
\label{sec:exp_complexity_analysis}
To understand how \method adapts its decomposition strategy to different complexity levels, we answer 3 research questions (RQs).

\paragraph{RQ1: How do we objectively measure input complexity?}
We select three complementary metrics to comprehensively and objectively measure input complexity: (1) context length captures information volume; (2) domain difficulty reflects inherent reasoning challenges; and (3) perplexity quantifies language uncertainty~\citep{li2024can,li2024tapilot,jelinek1977perplexity}. As shown in Figure~\ref{fig:complexity_analysis}, these metrics provide a systematic way to evaluate how different LLMs adapt the decomposition strategies to varying complexity levels.

\paragraph{RQ2: Does decomposition behavior show some patterns across different complexity dimensions?}
Figure~\ref{fig:complexity_analysis}, we observe consistent adaptation patterns within all LLMs. For example, as for the context length shown in Figure~\ref{fig:complexity_analysis}(a), the decomposition rate increases dramatically from 15\% (0-100 tokens) to 95\% (400+ tokens). All three complexity dimensions exhibit similar trends, where higher complexity consistently triggers more frequent decomposition. This consistency demonstrates the ability of \method to effectively detect complexity and dynamically adjust granularity via decomposition to reduce complexity.

\paragraph{RQ3: Do different LLMs share the same understanding of complexity?}
The results in Figure~\ref{fig:complexity_analysis} show that GPT-4o-mini constantly calls decomposition action more frequently across all complexity dimensions, revealing different complexity tolerance between GPT-4o and GPT-4o-mini, as discussed in Section~\ref{sec:adapt_granularity}. Rather than requiring manual complexity adjustments for each LLM, \method automatically perceives the complexity and dynamically adapts for different LLMs. This adaptive behavior enables robust performance across different LLMs without model-specific tuning. \revise{For example, as shown in Table~\ref{tab:overall_results}, although LLaMA-3.1-8B is smaller in size and less capable than LLaMA-3.1-70B, \method can still maintain robust performance via more decompose actions to adjust complexity, compared with GKP's deep performance drop. More analysis is in Appendix~\ref{app: more_complex_ana}.}

\begin{figure}[h]
    \centering
    \includegraphics[width=0.48\textwidth]{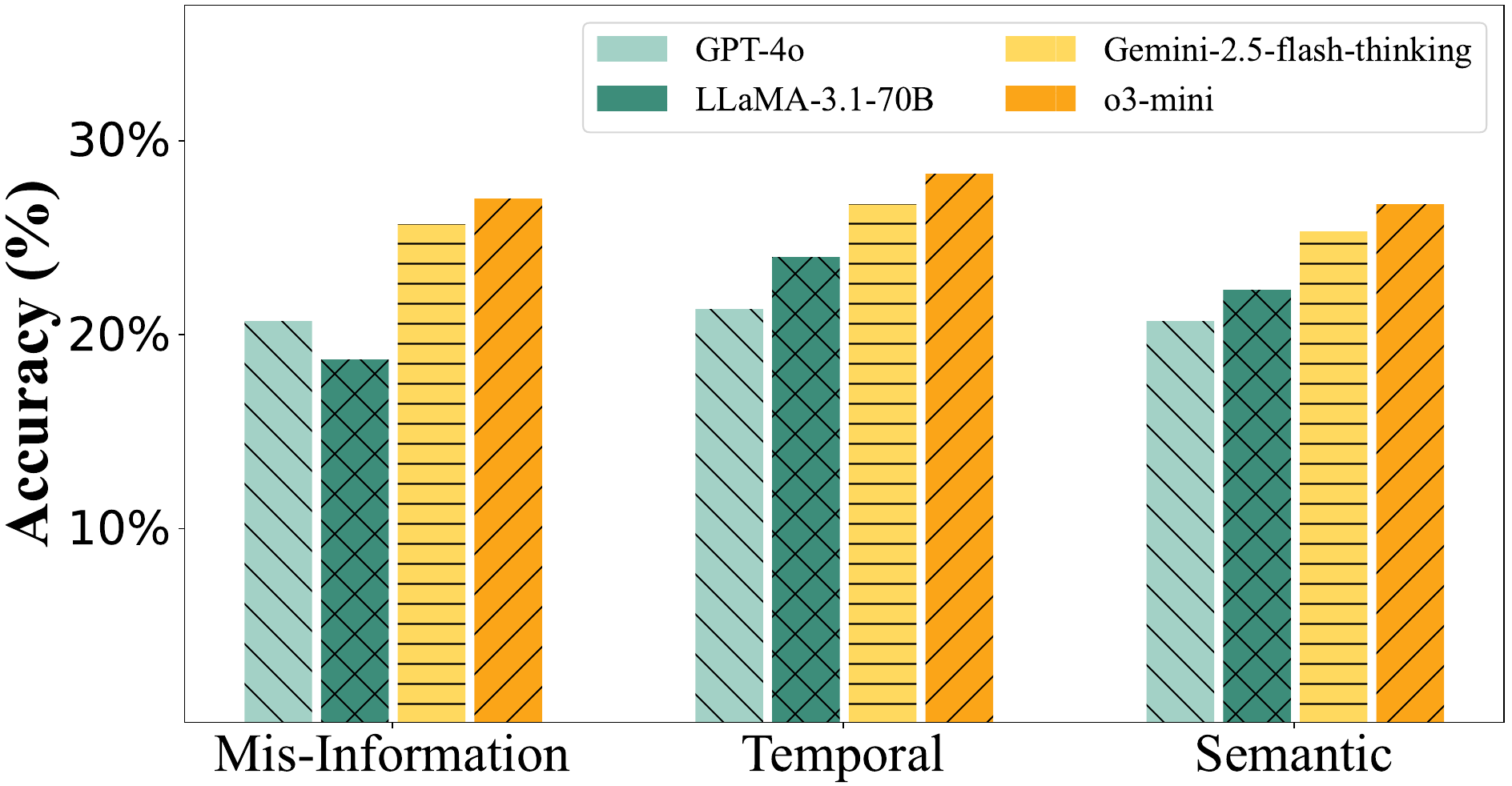}
    \caption{The performance of~\method using general LLMs and reasoning LLMs on 300 randomly sampled data for each conflict type.}
    \label{fig:reasoning}
    \vspace{-0.1cm}
\end{figure}

\subsection{General LLMs vs. Reasoning LLMs}

As illustrated in Figure~\ref{fig:reasoning}, the general-purpose LLMs (GPT-4o and Llama-3.1-70B) cluster together and attain comparatively low scores on all three conflict categories. In contrast, the reasoning-oriented models, Gemini-2.5-flash-thinking and o3-mini, form the top tier and consistently outperform the general models. The gap is most pronounced for misinformation conflicts, which are more amenable to reasoning-based resolution. For temporal and semantic conflicts, the gap narrows because over-rationalization issues arises more often, as discussed in detail in Section~\ref{sec: over}. In summary, stronger reasoning capability markedly boosts the performance of~\method. Although it also increases susceptibility to over-rationalization, \method can effectively mitigate this issue and still surpasses the general models.

\begin{table}[h]  
    \centering
    \resizebox{1.0\hsize}{!}{
    \begin{tabular}{lccc}  
    \toprule
    \textbf{\textsc{Method}} & \textbf{\textsc{Mis-Info.}} & \textbf{\textsc{Temporal}} & \textbf{\textsc{Semantic}} \\ 
    \midrule
    \method & 26.1 & 27.9 & 24.9\\
    w/o Navigational Actions & 18.4 \textbf{ ($\downarrow$ 7.7)} & 18.5 \textbf{ ($\downarrow$ 9.4)} & 15.7 \textbf{ ($\downarrow$ 9.2)} \\
    w/o Functional Actions & 13.8 \textbf{ ($\downarrow$ 12.3)} & 15.2 \textbf{ ($\downarrow$ 12.7)} & 13.3 \textbf{ ($\downarrow$ 11.6)} \\
    w/o DECOMPOSE Action  & 4.2 \textbf{ ($\downarrow$ 21.9)} & 4.5 \textbf{ ($\downarrow$ 23.4)} & 0.8 \textbf{ ($\downarrow$ 24.1)}\\
    \bottomrule
    \end{tabular}}
    \caption{Ablation study of \method in three datasets (conflict types) of ConflictBank. The numbers represent the accuracy in percentage. $\downarrow$ is an absolute decrease.}
    \label{tab:ablation}
\end{table}

\begin{figure*}[t]
    \centering
    \includegraphics[width=0.9\textwidth]{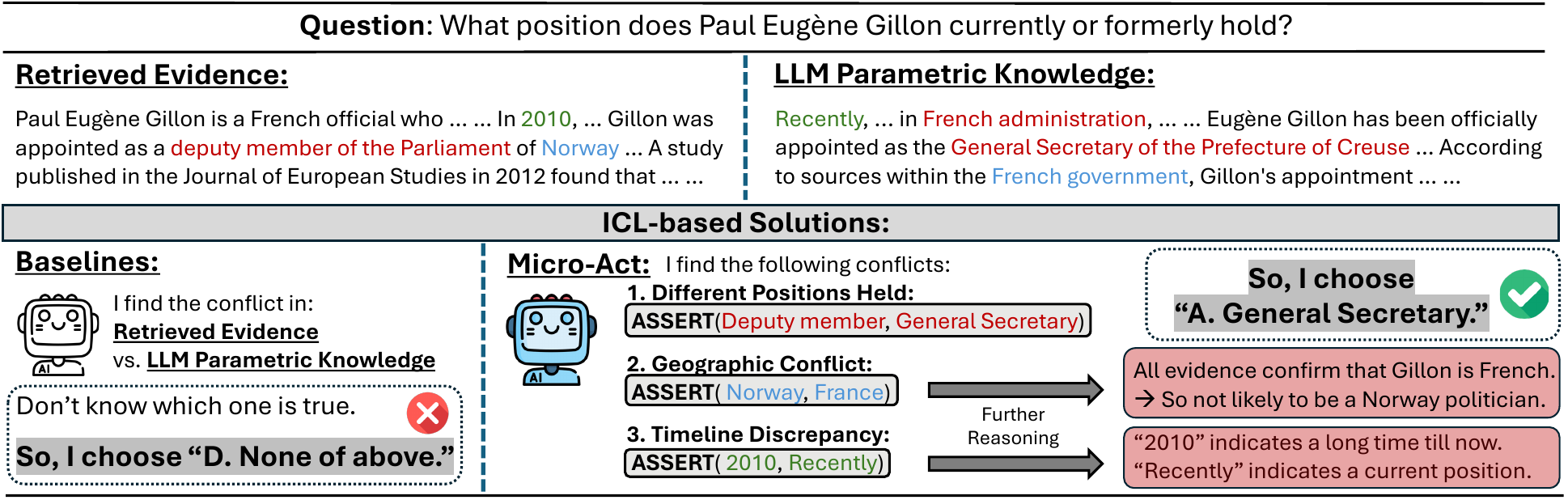}
    \caption{A case study of~\method and baselines models under a real knowledge conflict case.~\method can pinpoint fine-grained conflict points instead of being distracted by irrelevant context.}
    \label{fig:case_study}
\end{figure*}

\subsection{Ablation Study}
\label{sec:ablation_study}

Table~\ref{tab:ablation} presents an ablation study across all conflict types of ConflictBank.
Navigational and Functional actions serve as essential building blocks for conflict resolution, with their removal causing significant performance drops (9.4\% for navigational and 12.7\% for functional actions). While these actions are necessary for basic operations like context navigation and knowledge comparison, their effectiveness heavily depends on appropriate input granularity. Without proper guidance on the input granularity of those actions, the model struggles to maintain consistent performance, especially with complex contexts.

Decomposition action dramatically improves performance by dynamically adjusting input granularity for other actions. Its removal causes the most severe degradation (over 20\% performance drop), highlighting its crucial role. Through iterative decomposition, \method continuously refines inputs of other actions until they find the optimal granularity level where navigational and functional actions can operate most effectively. As discussed in Section~\ref{sec:adapt_granularity}, \method effectively elicits the latent ability of LLMs to perceive complexity and adapt to different environments. This adaptive process enhances the confidence of \method by ensuring all action components receive fine-grained information aligned with its capability, leading to higher accuracy in complex cases.

\subsection{Case Study} \label{sec: case_study}
\revise{In this case study, we demonstrate how~\method identifies nuanced conflicts underneath the superficial meaning of context which can hardly be located by simple side-by-side comparisons that baselines use. Consider the question ``\textit{What position does Paul Eugène Gillon currently or formerly hold?}'', where the retrieved context conflicts with LLM parametric knowledge as shown in Figure~\ref{fig:case_study}. \method can identify the different time references (\texttt{2010} vs. \texttt{recently}) and location (\texttt{Norway} vs. \texttt{France}). Then apply step-by-step reasoning to find the \textbf{\textit{underlying}} conflicts \textbf{\textit{beyond}} the \textbf{\textit{superficial}} context: (1) Majority consensus suggests he is from \texttt{France}, not \texttt{Norway}. (2) \texttt{2010} indicates a very long appointment, which is less likely compared with a \texttt{recent} appointment, given the question: \texttt{What position does Paul Eug\'ene Gillon currently or formerly hold?}
(3) And finally, determine that \texttt{Paul Eug\'ene Gillon} was \texttt{recently} appointed as a \texttt{French} politician then answer the question correctly.}

\section{Cost Analysis}
\method incurs a modest computational cost, where on ConflictBank it processes roughly 2.8 times more input tokens and 1.3 times more output tokens than the strongest baseline (GKP), translating to only \$0.008 extra per GPT-4o query and \$0.0005 with GPT-4o-mini, while inference latency rises by 0.6 s and 0.3 s respectively. Crucially, these overheads appear only when genuine conflicts trigger deeper decomposition; conflict-free questions finish as quickly as the baseline. Given the substantial gains in conflict-resolution accuracy reported in Table~\ref{tab:overall_results}, the marginal cost and delay are acceptable for real-world RAG deployments. Detailed token, cost, and timing breakdowns are provided in Appendix~\ref{app: cost}.

\section{Conclusion}
We proposed \textbf{\method}, a framework that addresses knowledge conflicts in RAG systems through hierarchical action decomposition. By automatically perceiving context complexity and breaking down comparisons into fine-grained steps, \method overcomes the limitations of simple side-by-side comparisons for example the \emph{over-rationalization} issue. Extensive experiments demonstrate its effectiveness across multiple datasets and conflict types, while maintaining strong robustness in non-conflict scenarios, making it particularly valuable for real-world RAG.

\newpage
\section{Limitations}
\label{sec:limitation}

While \method demonstrates strong performance in knowledge conflict resolution, several limitations warrant discussion. \revise{First, our \method needs additional intermediate steps to effectively pinpoint the conflicts underneath the superficial meaning of context, which can hardly be located by simple side-by-side comparisons that baselines use. Although baselines like end-to-end and COT~\citep{wei2022chain} are lightweight, their poor performance in knowledge conflict harms the effectiveness of RAG systems. We believe the efficiency should be considered after good performance. As detailed analyzed in Appendix~\ref{app: cost}, the extra overhead is relatively small and our analysis demonstrates that \method’s modest overhead is justified by its significantly enhanced conflict resolution performance.}
Second, our current evaluation focuses primarily on English language contexts. The effectiveness of decomposition strategies might vary across different languages and cultural contexts. 

Nevertheless, our work represents an important milestone in knowledge conflict resolution, establishing a strong foundation for future research in this critical area.

\section{Acknowledgement}
Reynold Cheng, Nan Huo, Jinyang Li, and Ge Qu are supported by the Hong Kong Jockey Club Charities Trust (Project 260920140), the University of Hong Kong (Project 2409100399), the HKU Outstanding Research Student Supervisor Award 2022-23, and the HKU Faculty Exchange Award 2024 (Faculty of Engineering). Bowen Qin was supported by National Science and Technology Major Project (Project 2022ZD0116306). Chenhao Ma was partially supported by NSFC under Grant 62302421, Basic and Applied Basic Research Fund in Guangdong Province under Grant 2023A1515011280, 2025A1515010439, Ant Group through CCF-Ant Research Fund, Shenzhen Research Institute of Big
Data under grant SIF20240004, and the Guangdong Provincial Key Laboratory of Big Data Computing, The Chinese University of Hong Kong, Shenzhen.

\section{Ethical Statement}
\label{sec:ethical_statement}

We prioritize ethical considerations throughout our research process. During data collection and preprocessing, we carefully filtered out examples containing sensitive, biased, or potentially harmful content to ensure our evaluation focuses on constructive knowledge resolution scenarios. Our in-context learning approach requires no additional training of language models, significantly reducing the environmental impact compared to fine-tuning methods. This aligns with growing concerns about the carbon footprint of AI research. Furthermore, all datasets used in this work are publicly available, ensuring reproducibility. 

\bibliography{acl_latex}

\clearpage
\appendix

\begin{table*}[h]
    \centering
    \resizebox{\linewidth}{!}{
    \begin{tabular}{lccccc}
        \toprule
        Method & Avg. \# Turns & Avg. Input Tokens & Avg. Output Tokens & Avg. Cost (USD) & Avg. Inference Time (s) \\
        \midrule
        COT & 2.0 & 652 & 421 & \$0.006 & 0.9 \\
        GKP & 2.0 & 1182 & 856 & \$0.012 & 1.3 \\
        \method & 3.4 & 3345 & 1137 & \$0.020 & 1.9 \\
        \bottomrule
    \end{tabular}}
    \caption{Cost analysis on GPT-4o.}
    \label{tab:gpt4o_cost}
\end{table*}

\begin{table*}[h]
    \centering
    \resizebox{\linewidth}{!}{
    \begin{tabular}{lccccc}
        \toprule
        Method & Avg. \# Turns & Avg. Input Tokens & Avg. Output Tokens & Avg. Cost (USD) & Avg. Inference Time (s) \\
        \midrule
        COT & 2.0 & 689 & 469 & \$0.0004 & 0.3 \\
        GKP & 2.0 & 1239 & 894 & \$0.0007 & 0.4 \\
        \method & 3.6 & 3532 & 1289 & \$0.0013 & 0.7 \\
        \bottomrule
    \end{tabular}}
    \caption{Cost analysis on GPT-4o-mini.}
    \label{tab:gpt4o_mini_cost}
\end{table*}

\section{\revise{Cost-Performance Trade-Off}}
\label{app: cost}
In this section, we provide a detailed cost and latency analysis on the ConflictBank dataset, comparing \method with the most powerful baseline within Generic Reasoning and Generation-aided Reasoning groups (i.e., COT and GKP) using GPT-4o and GPT-4o-mini, as shown in Table~\ref{tab:gpt4o_cost} and Table~\ref{tab:gpt4o_mini_cost}.

\subsection{Token Usage and Monetary Cost}

\begin{itemize}
    \item \textbf{Input Tokens:} Because \method dynamically decomposes conflicts, it initiates additional turns to resolve contradictions. As a result, it uses about 2.8× more input tokens than GKP.
    \item \textbf{Output Tokens:} \method's output length is roughly 1.3× higher than GKP. Although it writes multiple short intermediate responses, the final output does not explode in length, as the output of each intermediate step tends to be relatively concise.
    \item \textbf{Overall Cost:} The cost difference comes to an additional \$0.008 per query for GPT-4o. For GPT-4o-mini, the extra cost is only \$0.0005 per query, which is fairly small.
\end{itemize}

\subsection{Inference Time Overhead}

\begin{itemize}
    \item \textbf{GPT-4o:} \method takes 1.9s on average, which is about 0.6s longer than GKP (1.3s).
    \item \textbf{GPT-4o-mini:} \method requires 0.7s on average, 0.3s longer than GKP (0.4s).
    \item The latency shown above is not using multi-threading. If we use multi-threading, the extra inference latency will be further reduced significantly.
\end{itemize}

This additional overhead comes from the extra decomposition steps in scenarios where conflicts are actually perceived by \method. However, for conflict-free queries, \method performs fewer steps, avoiding this overhead.

\subsection{Justification of Additional Overhead}

\begin{enumerate}
    \item \textbf{Significant Performance Gain:} As shown in our main experiments, \method achieves \textbf{notable improvements} in resolving knowledge conflicts. It indicates that our dynamic decomposition approach is essential for detecting finer-grained conflicts.
    \item \textbf{Adaptive Depth:} \method only needs deeper decomposition when a conflict is perceived, which is necessary to locate the underlying conflicts that baselines cannot find. On conflict-free questions, it quickly finalizes the answer, keeping cost and latency low.
    \item \textbf{Practical Applicability:} In many real-world applications such as SWE-agent \citep{yang2024swe}, baseline query costs more than \$2 per query. We believe that an additional \$0.008 in GPT-4o (and \$0.0005 for GPT-4o-mini) and 0.6 extra seconds in GPT-4o (and 0.3 sec for GPT-4o-mini) per query is acceptable given the significantly improved performance, especially for real-world scenarios.
\end{enumerate}

We believe this analysis demonstrates that \method's \textbf{modest overhead} is justified by its \textbf{significantly enhanced} conflict resolution capabilities.

\begin{figure}[h]
    \centering
    \includegraphics[width=0.48\textwidth]{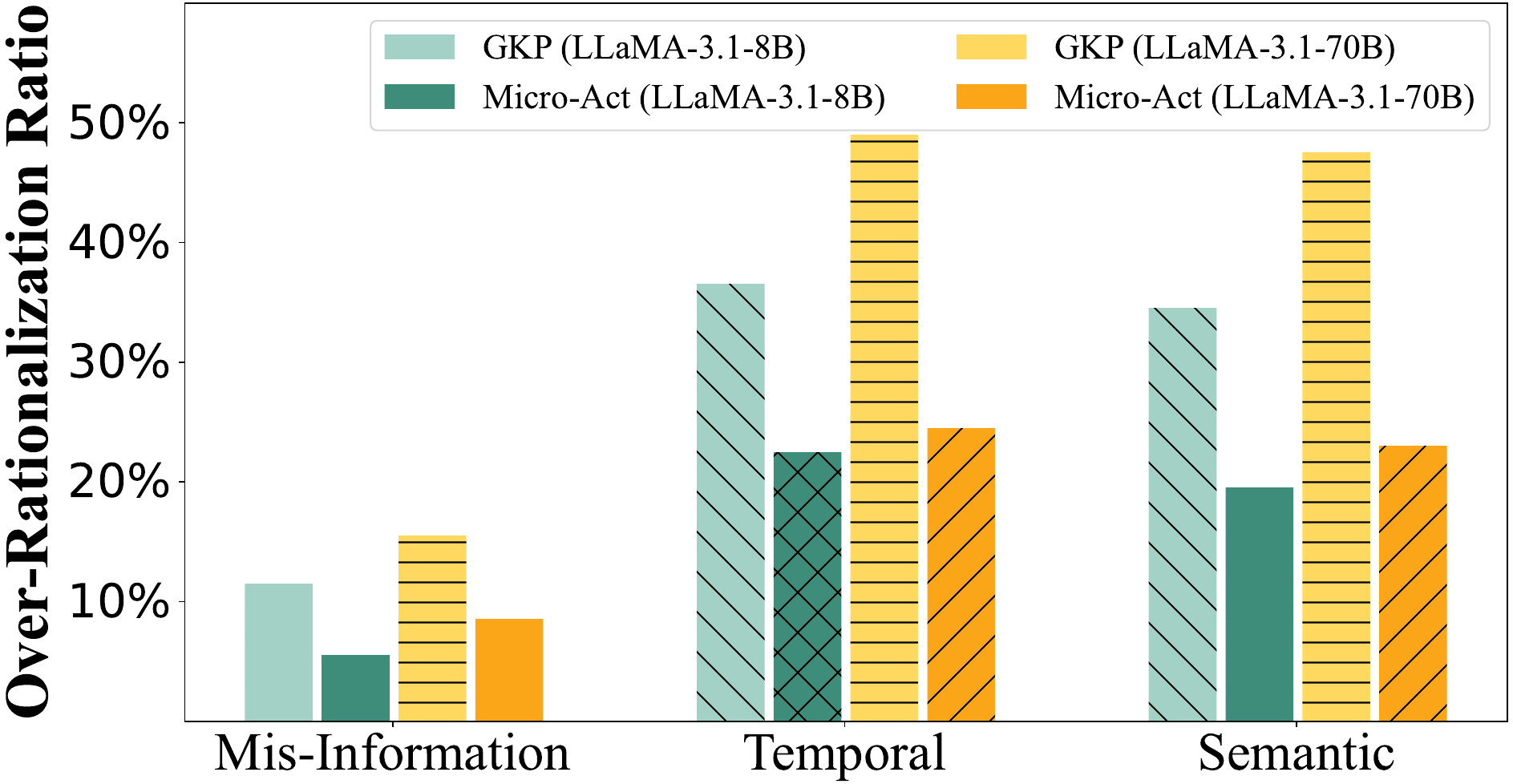}
    \caption{The Over-Rationalization Ratio of GKP and \method using LLaMA-3.1-8B and LLaMA-3.1-70B (lower is better).}
    \label{fig:over_rate}
\end{figure}

\section{\revise{Over-Rationalization Issue and Analysis}}
\label{app: Over-Rationalization}

In this section, we present a quantitative investigation on 600 queries (200 from each conflict type) in the ConflictBank dataset, comparing GKP (strongest baseline) and \method on two LLMs: LLaMA-3.1-70B and GPT-4o-mini. We measure the proportion of queries in which the model exhibits \textit{over-rationalization} (i.e., rationalizing contradictory facts in their step-by-step chain-of-thought reasoning).

From Figure~\ref{fig:over_rate}, it is clear that all models exhibit a higher ratio of over-rationalization under temporal and semantic conflict types, as these two types are more easily rationalized. For example, as shown in Figure~\ref{fig:case_study}, conflicting knowledge about whether \texttt{Paul Eug\'ene Gillon} was appointed in \texttt{2010} or more \texttt{recently} can both be reasonable, making it difficult for the model to rely on retrieved evidence or LLM parametric knowledge.

The GKP method shows a higher tendency to rationalize these conflicts, as LLMs are easily distracted by complex context \citep{GSMSymbolic}. This leads to failure in identifying the fine-grained conflicts underneath. In contrast, \method dynamically decomposes the complex context to reduce its complexity and identifies the conflicts behind the superficial context. This enables \method to pinpoint conflict points and reason on them for correct answers.

\subsection{A More Detailed Case Study}

In Figure~\ref{fig:case_study}, an example of contradictory information regarding \texttt{Paul Eug\'ene Gillon} being appointed in \texttt{2010} versus a more \texttt{recent} appointment is shown. Additionally, incorrect evidence suggests he might be a politician in \texttt{Norway}. By iteratively decomposing the knowledge, \method is able to:

\begin{enumerate}
    \item Identify the conflicting time references (\texttt{2010} vs. \texttt{recently}) and location (\texttt{Norway} vs. \texttt{France}).
    \item Apply step-by-step reasoning to find the underlying conflicts beyond the superficial context:
    \begin{itemize}
        \item Majority consensus suggests he is from \texttt{France}, not \texttt{Norway}.
        \item \texttt{2010} indicates a very long appointment, which is less likely compared with a \texttt{recent} appointment, given the question: \texttt{What position does Paul Eug\'ene Gillon currently or formerly hold?}
    \end{itemize}
    \item Finally, determine that \texttt{Paul Eug\'ene Gillon} was \texttt{recently} appointed as a \texttt{French} politician and correctly answer the question.
\end{enumerate}

These nuanced conflicts are hidden beneath the superficial meaning of context and can hardly be detected by simple side-by-side comparisons used by baseline models.

\section{Perplexity Calculation}
\label{sec:perp_calc}

In this section, we provide a detailed explanation of how to compute the perplexity (PPL) of a given text using the GPT-2 language model, which is not any of the language models used in our work. The aim is to provide an objective measurement of knowledge context complexity. Perplexity is a widely used metric for evaluating the quality of language models, indicating how well the model predicts a sample of text~\citep{jelinek1977perplexity}. In this work, by fixing the language model to be GPT-2, we use PPL to measure the complexity of context text. A lower perplexity value generally corresponds to a lower context text complexity.

\subsection{Formal Definition of Perplexity}

Let the text be represented as a sequence of tokens:
\begin{equation}
W = w_1, w_2, \ldots, w_N,
\end{equation}
where each \( w_i \) is a token (e.g., a subword unit as utilized by GPT-2).

Given a language model that estimates the probability of each token conditioned on all previous tokens, the joint probability of the sequence \( W \) can be factorized as:
\begin{equation}
P(W) = \prod_{i=1}^{N} P(w_i \mid w_1, w_2, \ldots, w_{i-1}).
\end{equation}

The perplexity of the sequence \( W \) under the model is defined as:
\begin{equation}
\begin{aligned}
    &\text{PPL}(W) = \\
    &\exp\left(-\frac{1}{N}\sum_{i=1}^{N}\ln P(w_i \mid w_1, w_2, \ldots, w_{i-1})\right).
\end{aligned}
\end{equation}

In other terms, if we use base-2 logarithms:
\begin{equation}
\text{PPL}(W) = 2^{-\frac{1}{N}\sum_{i=1}^{N}\log_2 P(w_i \mid w_1, \ldots, w_{i-1})}.
\end{equation}

The perplexity can be interpreted as the effective average branching factor of the language model. A perfectly predicting model (one that assigns probability 1 to the observed sequence) would achieve a perplexity of 1.

\subsection{GPT-2-Based Computation}

GPT-2 is a Transformer-based language model trained on large-scale text data. It provides a probability distribution over the next token \( w_i \) given the previous tokens \((w_1, \ldots, w_{i-1})\). Formally, GPT-2 implements:
\begin{equation}
P(w_i \mid w_1, \ldots, w_{i-1}) = \text{softmax}(h_{i-1}W_e)_{w_i},
\end{equation}
where \( h_{i-1} \) is the hidden state vector produced by the Transformer after processing tokens \(w_1, \ldots, w_{i-1}\), and \( W_e \) is the token embedding matrix used to map hidden states to logits over the vocabulary. The \(\text{softmax}\) function converts these logits into probabilities.

\subsection{Practical Steps for Perplexity Calculation}

To compute perplexity using GPT-2 in practice, one would proceed as follows:

\paragraph{Tokenization.} 
   Convert the raw text into GPT-2 compatible tokens:
   \begin{equation}
   \text{Text} \xrightarrow{\text{tokenizer}} (w_1, w_2, \ldots, w_N).
   \end{equation}

\paragraph{Model Inference.}  
   For each token \( w_i \), feed the preceding tokens \((w_1, w_2, \ldots, w_{i-1})\) into GPT-2 to obtain:
   \begin{equation}
   P(w_i \mid w_1, \ldots, w_{i-1}).
   \end{equation}
   This is done by running the model $\mathcal{M}$ which is GPT-2 forward pass:
   \begin{equation}
   (h_1, h_2, \ldots, h_{i-1}) = \mathcal{M}(w_1, w_2, \ldots, w_{i-1})
   \end{equation}
   and then applying the softmax over the output logits to get the probability of \( w_i \).

\paragraph{Log Probability Computation.}  
   The next step is to extract \(\ln P(w_i \mid w_1, \ldots, w_{i-1})\) from the model’s output distribution.

\paragraph{Summation and Normalization.}  
   Compute the average negative log-probability:
   \begin{equation}
   -\frac{1}{N}\sum_{i=1}^{N}\ln P(w_i \mid w_1, \ldots, w_{i-1}).
   \end{equation}

\paragraph{Exponentiation}  
   Take the exponential of this value to obtain the perplexity:
   \begin{equation}
   \begin{aligned}
       &\text{PPL}(W) = \\
       &\exp\left(-\frac{1}{N}\sum_{i=1}^{N}\ln P(w_i \mid w_1, \ldots, w_{i-1})\right).
   \end{aligned}
   \end{equation}

\section{Mathematical Derivation of Knowledge Representation Transitions}
\label{sec:appendix_derivation}

In this section, we present a formal derivation of the knowledge representation transition process. We begin by defining the key probability distributions:

\begin{itemize}
\item $p_t(K_n)$: The distribution of knowledge representation at step $t$
\item $p_{model}(K_n|K)$: The probability of the model generating new knowledge representation $K_n$ from previous state $K$
\item $p_{verify}(c|K_n)$: The probability of the verifier generating conflict detection result $c$ based on knowledge representation $K_n$
\end{itemize}

Following the principles of probabilistic state transitions, we can establish two fundamental equations:

\subsection{Knowledge Representation Transition Equation}
The transition to new knowledge representation can be expressed as:

\begin{equation}
p_t(K_n) = \sum_{c'} p_{model}(K_n|c') \cdot p_{t-1}(c')
\end{equation}

This equation represents how new knowledge representation are derived from previous conflict detection results.

\subsection{Conflict Detection Equation}
The probability distribution of conflict detection results is given by:

\begin{equation}
p_{t-1}(c') = \sum_{K} p_{verify}(c'|K) \cdot p_{t-1}(K)
\end{equation}

This captures how conflict detection results are generated based on previous knowledge representation.

\subsection{Combined Transition Model}
Substituting the conflict detection equation into the state transition equation yields:

\begin{equation}
\begin{aligned}
p_t(K_n) = &\sum_{c'} p_{model}(K_n|c') \\
&\cdot [\sum_{K} p_{verify}(c'|K) \cdot p_{t-1}(K)]
\end{aligned}
\end{equation}

Rearranging the summation order:

\begin{equation}
\begin{aligned}
p_t(K_n) = &\sum_{c'} \sum_{K} p_{model}(K_n|c')\\
&\cdot p_{verify}(c'|K) \cdot p_{t-1}(K)
\end{aligned}
\end{equation}

\subsection{Bayesian Formulation}
Applying Bayes' rule to transform $p_{verify}(c'|K)$ into $p_{verify}(K|c')$:

\begin{equation}
\begin{aligned}
p_t(K_n) = &\sum_{K,c'} p_{t-1}(K)\\
&\cdot p_{verify}(K|c') \cdot p_{model}(K_n|c')
\end{aligned}
\end{equation}

This final form encapsulates three key components:
\begin{itemize}
\item $p_{t-1}(K)$: Distribution of previous knowledge representation
\item $p_{verify}(K|c')$: Verifier's evaluation of knowledge
\item $p_{model}(K_n|c')$: Model's probability of generating new knowledge based on conflict detection
\end{itemize}

This formulation provides a comprehensive mathematical framework for analyzing the evolution of knowledge representation through iterative refinement and verification.

\begin{figure*}[h]
    \centering
    \includegraphics[width=0.98\textwidth]{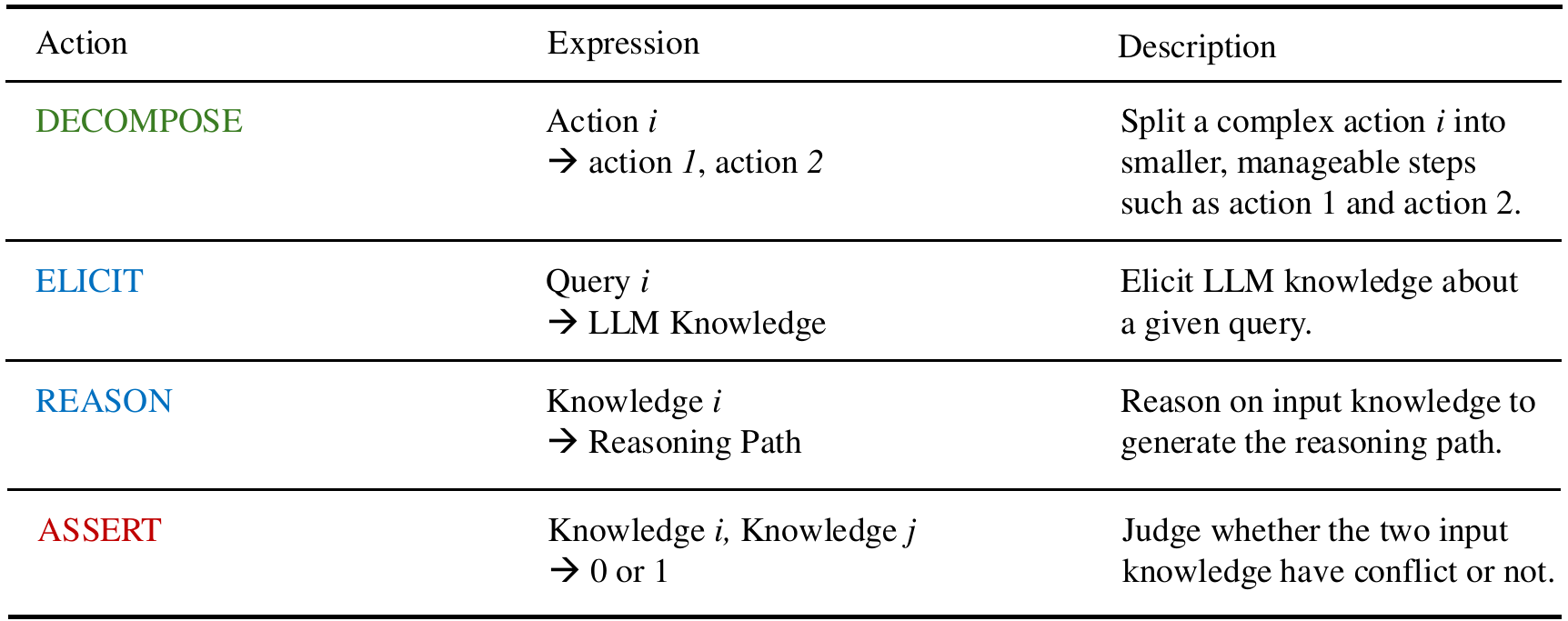}
    \caption{The table of actions in our hierarchical action space.}
    \label{fig:actions}
\end{figure*}

\section{Error Analysis}
\label{sec:error_analysis}

To better understand the limitations of \method, we conducted a detailed error analysis on 1,000 randomly sampled examples across all five datasets. Our analysis revealed two predominant error types:

\paragraph{Context Distraction (63\% of errors)}
Despite our decomposition strategy, LLMs occasionally become overwhelmed by complex contexts and default to expressing uncertainty ("I don't know" or "Cannot determine"). This typically occurs when the context contains multiple interrelated facts or complex logical relationships that challenge the model's ability to maintain coherent reasoning chains. For instance, in multi-hop reasoning questions where evidence pieces are densely connected, even decomposed segments may retain inherent complexity that exceeds the model's processing capacity.

\paragraph{Over-reliance on Retrieved Evidence (37\% of errors)}
The second major error type manifests when LLMs exhibit a strong bias toward retrieved evidence, even when it conflicts with their parametric knowledge. This behavior is particularly prominent in cases where the retrieved evidence appears more specific or detailed than the model's inherent knowledge. Such errors suggest that while \method effectively identifies conflicts, the final resolution step may still be influenced by an implicit bias toward explicit external information over learned knowledge.

These findings indicate that while \method significantly improves conflict resolution, future work should focus on enhancing the model's ability to maintain reasoning clarity in highly complex scenarios and developing more balanced weighing mechanisms between retrieved and parametric knowledge.

\section{\revise{More Complexity Aspect Analysis}} \label{app: more_complex_ana}

Besides the complexity aspects discussed in Figure~\ref{fig:complexity_analysis}, we explore how the number of decomposition steps (i.e., the number of times \method invokes its ``DECOMPOSE'' action) varies across different conflict types. We collect more results on temporal, misinformation, and semantic conflicts for both GPT-4o and GPT-4o-mini. Below is a summary of the average number of decomposition steps taken for each conflict category:

\begin{table}[h]
    \centering
    \caption{Average DECOMPOSE action steps per conflict type.}
    \resizebox{1.0\hsize}{!}{
    \begin{tabular}{lcc}
        \toprule
        \textbf{Conflict Type} & \textbf{GPT-4o (Avg. Steps)} & \textbf{GPT-4o-mini (Avg. Steps)} \\
        \midrule
        Misinformation & 0.8 & 1.3 \\
        Temporal & 1.6 & 2.2 \\
        Semantic & 1.5 & 2.3 \\
        \bottomrule
    \end{tabular}}
\end{table}

\paragraph{Temporal and Semantic Conflicts.}
We can see that temporal and semantic conflict types usually trigger a higher number of decomposition steps. As detailed in Appendix~\ref{app: Over-Rationalization}, these two types are more easily to be \textbf{``over-rationalized''}. Thus, more decomposition steps are needed to investigate underneath conflicts and the logic flaws behind the superficial context, which looks reasonable.

\paragraph{Misinformation Conflicts.}
Misinformation typically involves more \textbf{superficial conflicts}, which is more intuitive. As a result, fewer DECOMPOSE actions are invoked because there is less ambiguity in the evidence to untangle.

\paragraph{Why Decompose More?}
The observed more decomposition is mainly because the nuanced conflicts are \textbf{underneath the superficial context} and can hardly be located by simple side-by-side comparisons that baselines use, detailed discussion can be found in Appendix~\ref{app: Over-Rationalization}.

\paragraph{Model Size \& Complexity.} 
We observe that GPT-4o-mini has a higher decomposition step count across all conflict types. Smaller LLMs often require additional steps to reduce complexity, revealing different complexity tolerance between GPT-4o and GPT-4o-mini, as discussed in Section~\ref{sec:adapt_granularity}. A more detailed discussion can be found in Section~\ref{sec:exp_complexity_analysis}, RQ3.

This illustrates exactly \textbf{why} and \textbf{how} \method's dynamic reasoning pipeline triggers additional decomposition for temporal and semantic conflicts, where potential ``over-rationalizations'' are more likely to arise.

\section{Implementation Details}
\label{imp_detail}
\subsection{Action Details}
All the actions designed in our proposed hierarchical action space are illustrated in Table~\ref{fig:actions}.

\subsection{Generic Reasoning Models}
\begin{figure*}[h]
    \centering
    \includegraphics[width=0.98\textwidth]{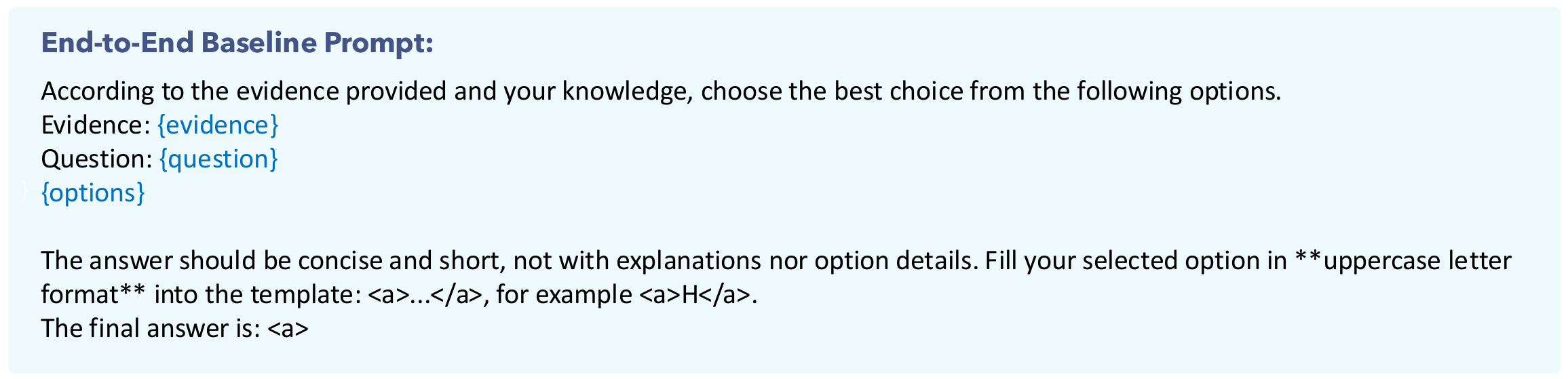}
    \caption{The prompt of the End-to-End QA baseline method.}
    \label{fig:end2end}
\end{figure*}

\paragraph{End-to-End QA.} The End-to-End QA prompt as shown in Figure~\ref{fig:end2end}, directly provides the model with the question and requests an immediate, self-contained answer. It contains no intermediate reasoning instructions, and the model is expected to return its best guess in a single generation pass. This approach assumes the model’s internal representations are sufficient for reasoning without explicitly prompting it to show work.

\begin{figure*}[h]
    \centering
    \includegraphics[width=0.98\textwidth]{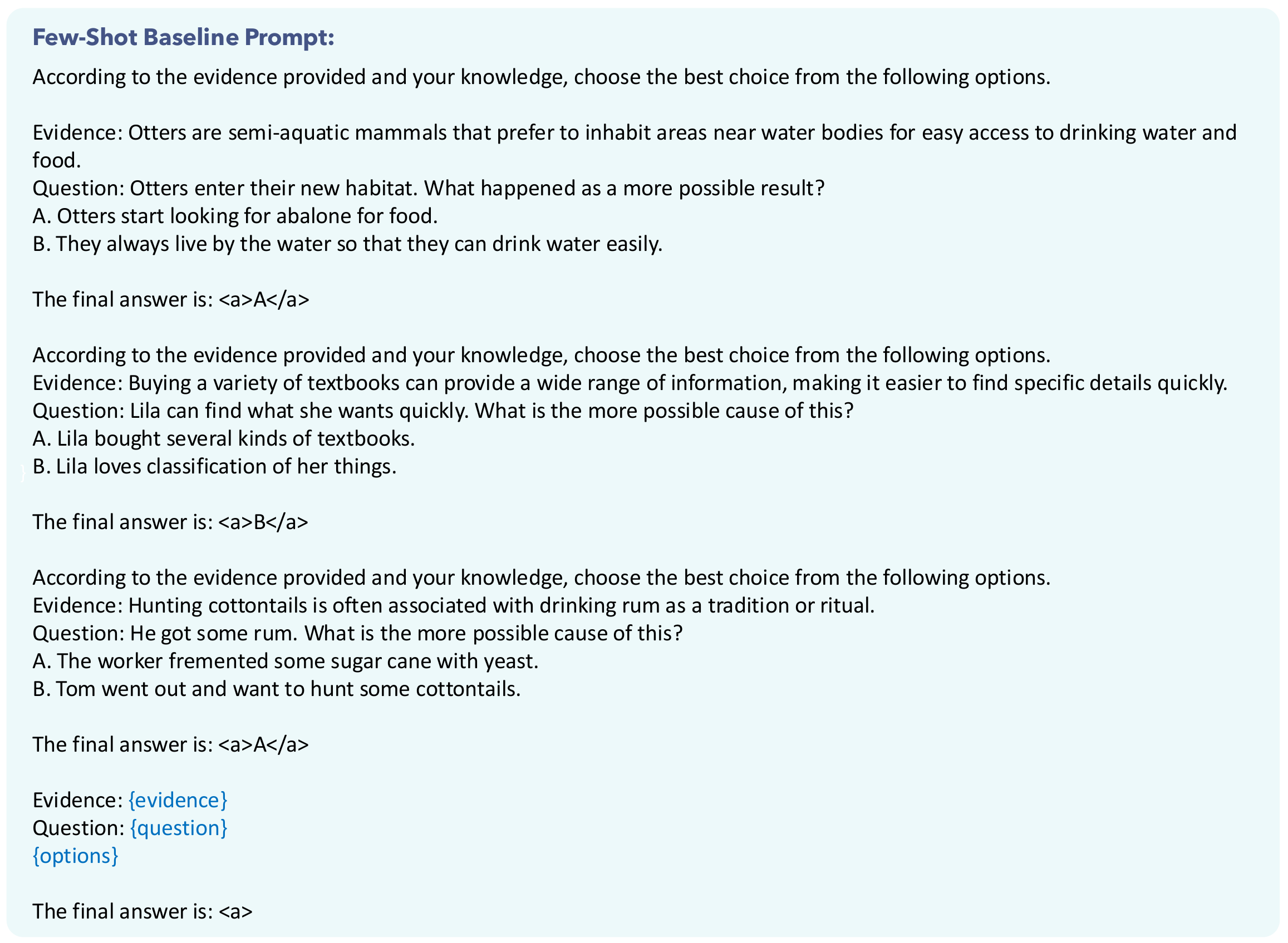}
    \caption{The prompt of the Few-Shot QA baseline method.}
    \label{fig:few_shot}
\end{figure*}

\paragraph{Few-Shot QA.} The Few-Shot QA prompt as shown in Figure~\ref{fig:few_shot}, includes one or more example QA pairs before presenting the target question. These examples help the model align its reasoning with the desired output format and style. The provided examples are chosen to be representative of the question domain and complexity level.

\begin{figure*}[h]
    \centering
    \includegraphics[width=0.98\textwidth]{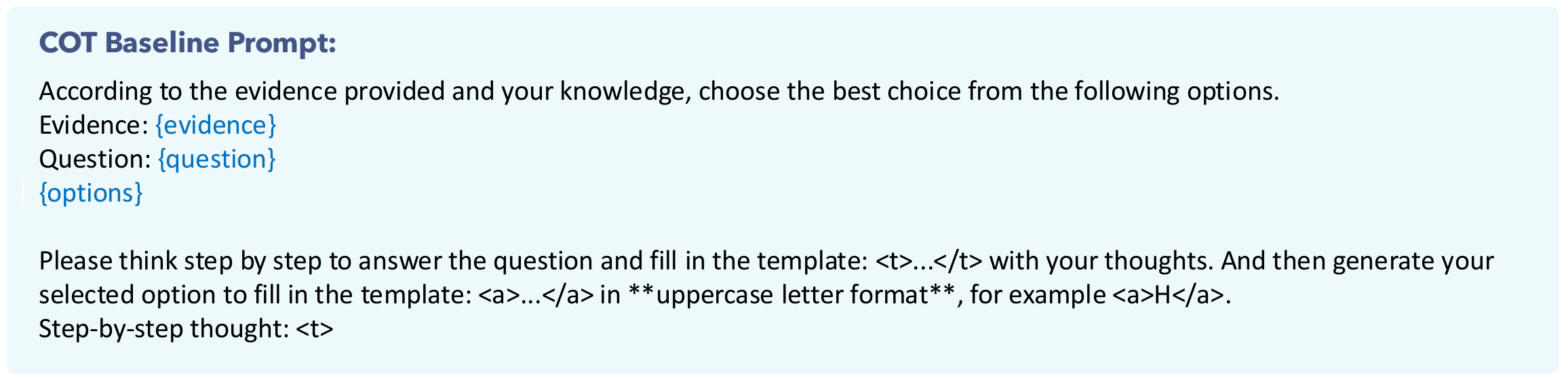}
    \caption{The prompt of the COT~\citep{wei2022chain} baseline method.}
    \label{fig:cot}
\end{figure*}

\paragraph{Chain-of-Thought.} The Chain-of-Thought~\citep{wei2022chain} prompt as shown in Figure~\ref{fig:cot}, instructs the model to show its intermediate reasoning steps explicitly. After presenting the question, the prompt requests the model to “think aloud” by outlining its reasoning process before concluding with a final, concise answer. This approach encourages the model to form more coherent and justifiable solutions.

\subsection{Generation-aided Reasoning Models}
\begin{figure*}[h]
    \centering
    \includegraphics[width=0.98\textwidth]{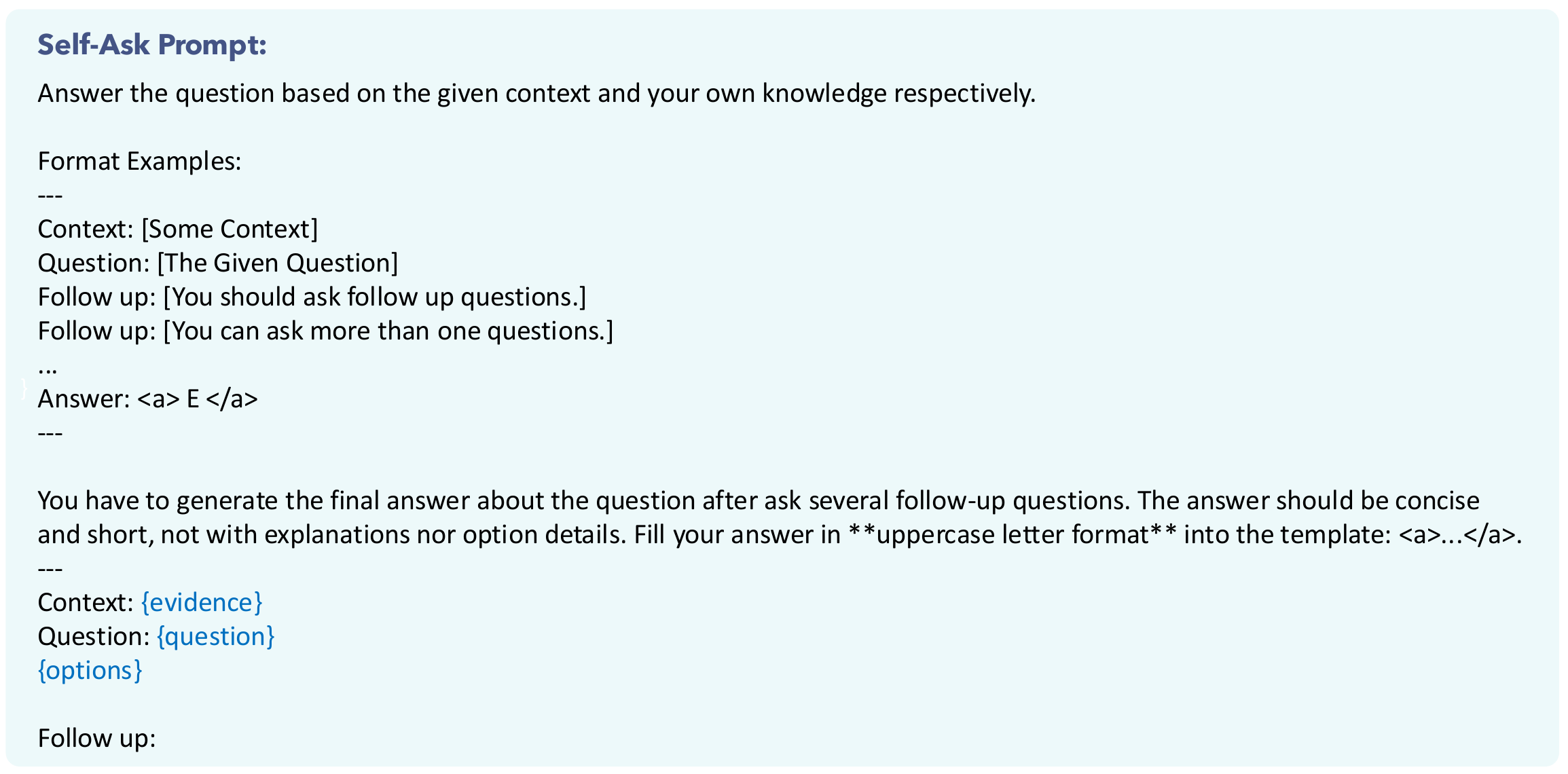}
    \caption{The prompt of the Self-Ask~\citep{press2023measuring} QA baseline method.}
    \label{fig:self_ask}
\end{figure*}

\paragraph{Self-Ask.} The Self-Ask~\citep{press2023measuring} prompt as shown in Figure~\ref{fig:self_ask}, breaks down a complex question into sub-questions and then prompts the model to answer them step-by-step. By iteratively generating and resolving subtasks, the model can handle multi-step reasoning tasks more systematically, ultimately consolidating the intermediate answers into a final solution.

\begin{figure*}[h]
    \centering
    \includegraphics[width=0.98\textwidth]{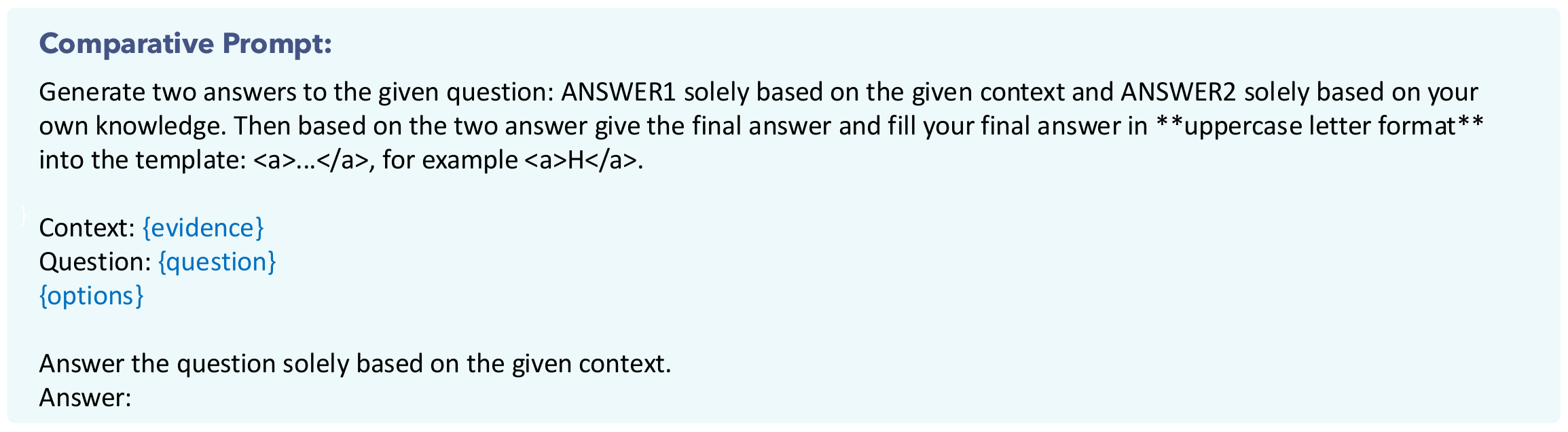}
    \caption{The prompt of the comparative~\citep{wang2023resolving} baseline method.}
    \label{fig:comparative}
\end{figure*}

\paragraph{Comparative QA.} The Comparative QA~\citep{wang2023resolving} prompt as shown in Figure~\ref{fig:comparative}, asks LLMs generate the answer of the question regardless of the retrieved evidence at first. Then answer the question by considering both the retrieved evidence and the self-generated answer.

\paragraph{Generation Phase of GKP.} In the Generation Phase of GKP~\citep{liu2022generated}, the prompt encourages the model to generate the knowledge needed to answer the given question. The model then lists relevant knowledge without yet providing the final answer. 

\paragraph{Answering Phase of GKP.} Once the self-generated knowledge is established, the Answering Phase of GKP~\citep{liu2022generated} prompt feeds the previously generated knowledge back into the model. Using this as a guide, the model now produces a final answer. This is a two-step process.

\subsection{\method}
\begin{figure*}[h]
    \centering
    \includegraphics[width=0.75\textwidth]{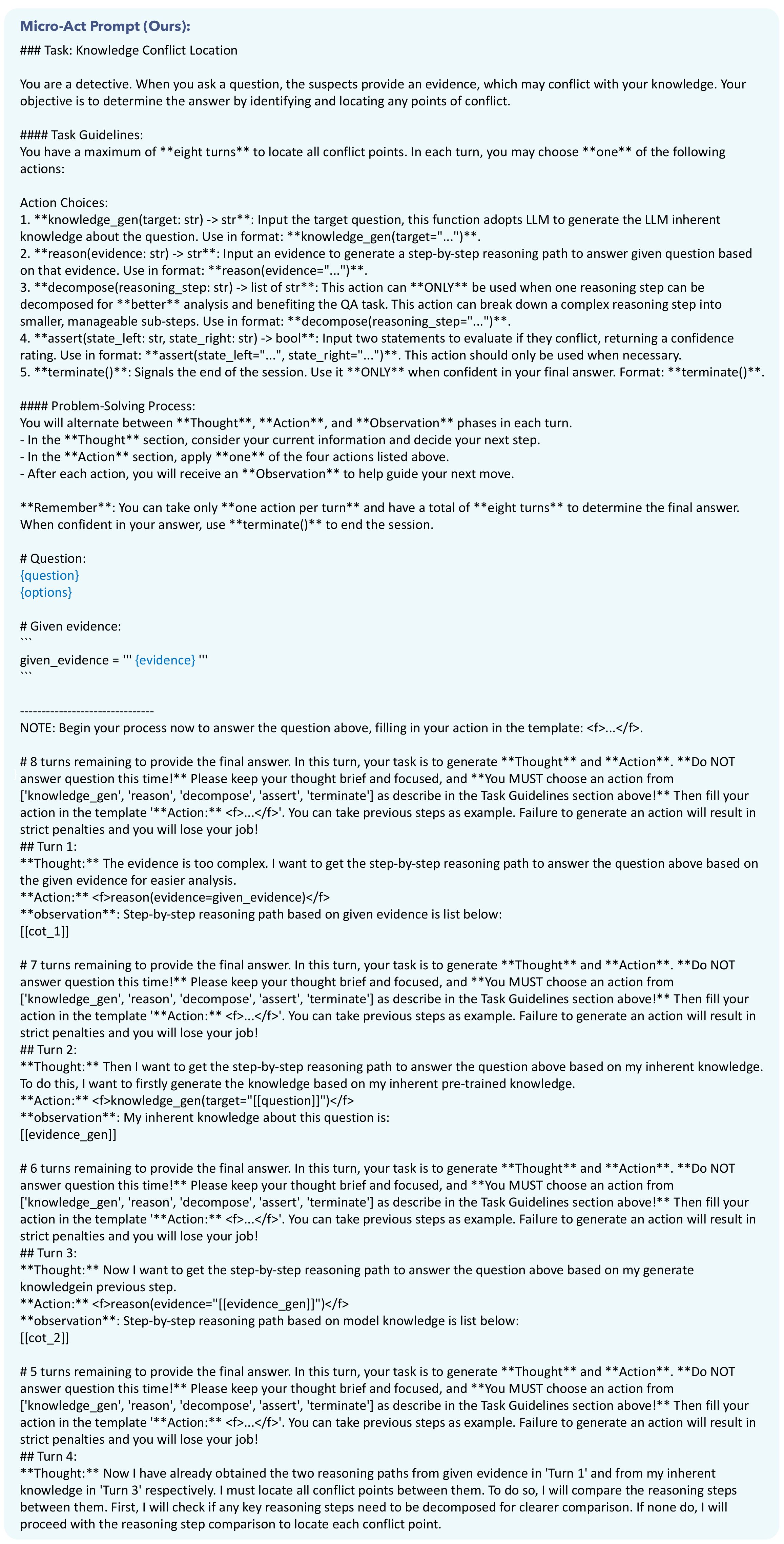}
    \caption{The prompt of our proposed \method method.}
    \label{fig:meta_act_prompt}
\end{figure*}

\paragraph{\method.} The prompt of our proposed \method model is shown in Figure~\ref{fig:meta_act_prompt}.

\section{Human Evaluation}
\label{human_eval}

To assess the quality and representativeness of our 1,000-instance samples, we conducted a human study with 10 volunteer expert annotators, each having substantial experience in QA tasks.

\paragraph{Evaluation Procedure.}
For each dataset, we presented each annotator with a total of 100 QA items: 50 randomly drawn from the full dataset and 50 randomly drawn from the 1,000-instance sample. Annotators were blind to which items came from which source. Each annotator answered all 100 questions to the best of their ability.

\paragraph{Measurements.}
We measured annotator accuracy, defined as the proportion of correct answers, on both subsets. Across all datasets, the average accuracy on sample-based QA pairs differed by less than 5\% from that on the corresponding full-dataset pairs. This consistency suggests that the sampled subsets do not introduce systematic bias in terms of difficulty or content distribution.

\paragraph{Internal Agreement.}
To ensure that results were not driven by a few outliers, we examined internal agreement among the 10 annotators. We computed Fleiss’ kappa~\citep{fleiss1971measuring}, which was consistently above 0.80 for all datasets, indicating substantial agreement. In addition, the standard deviation of accuracy across annotators remained under 2\% for each subset type, reflecting stable and consistent performance patterns.

These findings demonstrate that our sampling strategy preserves the key characteristics of the original datasets, maintaining both content diversity and difficulty level, and that our evaluation results are reliable and robust across multiple independent annotators. After the review period, we will open-source the sampled datasets for reproduction and for researchers to develop more advanced methods on knowledge conflict.

\section{Model Descriptions}
\label{sec:model_des}
Our empirical evaluation employs three representative language models, each positioned at different capability levels.

\paragraph{GPT-4o.}
GPT-4o~\citep{openai2023gpt4} is a state-of-the-art foundation model that excels at complex reasoning tasks. Our experiments leverage its robust instruction-following capabilities and advanced reasoning abilities to evaluate the upper bounds of adaptive complexity.

\paragraph{GPT-4o-mini.}
GPT-4o-mini~\citep{openai2023gpt4} is a balanced model that combines computational efficiency with strong reasoning capabilities. This model serves as an ideal testbed for examining how moderate model capacity influences the granularity of knowledge decomposition across varying task complexities.

\paragraph{LLaMA-3.1-70B-Instruct.}
LLaMA-3.1-70B-Instruct~\citep{Dubey2024TheL3} is a 70-billion parameter large language model built on the LLaMA architecture. This instruction-tuned variant exhibits strong performance across diverse NLP tasks, with particular strengths in reasoning and coherent text generation.

\paragraph{LLaMA-3.1-8B-Instruct.}
LLaMA-3.1-8B-Instruct~\citep{Dubey2024TheL3} is a 8-billion parameter large language model built on the LLaMA architecture. This instruction-tuned variant exhibits strong performance across diverse NLP tasks, with particular strengths in reasoning and coherent text generation.

\begin{figure*}[h]
    \centering
    \includegraphics[width=0.98\textwidth]{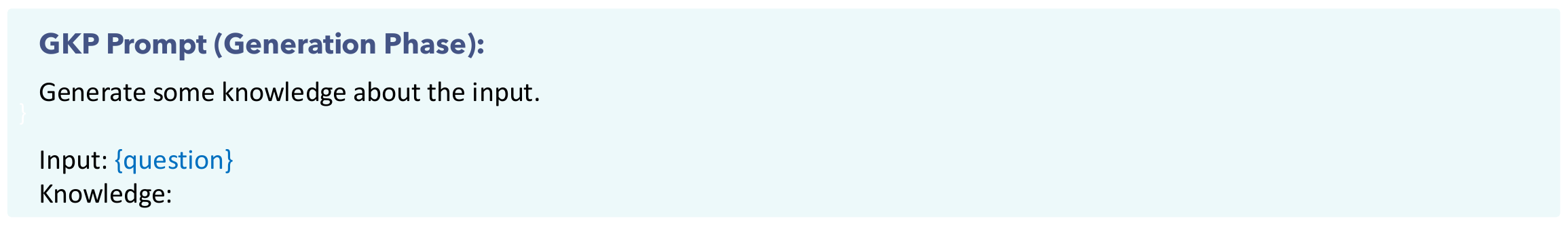}
    \caption{The prompt of the generation phase of GKP~\citep{liu2022generated} baseline method.}
    \label{fig:GKP_gen}
\end{figure*}

\begin{figure*}[h]
    \centering
    \includegraphics[width=0.98\textwidth]{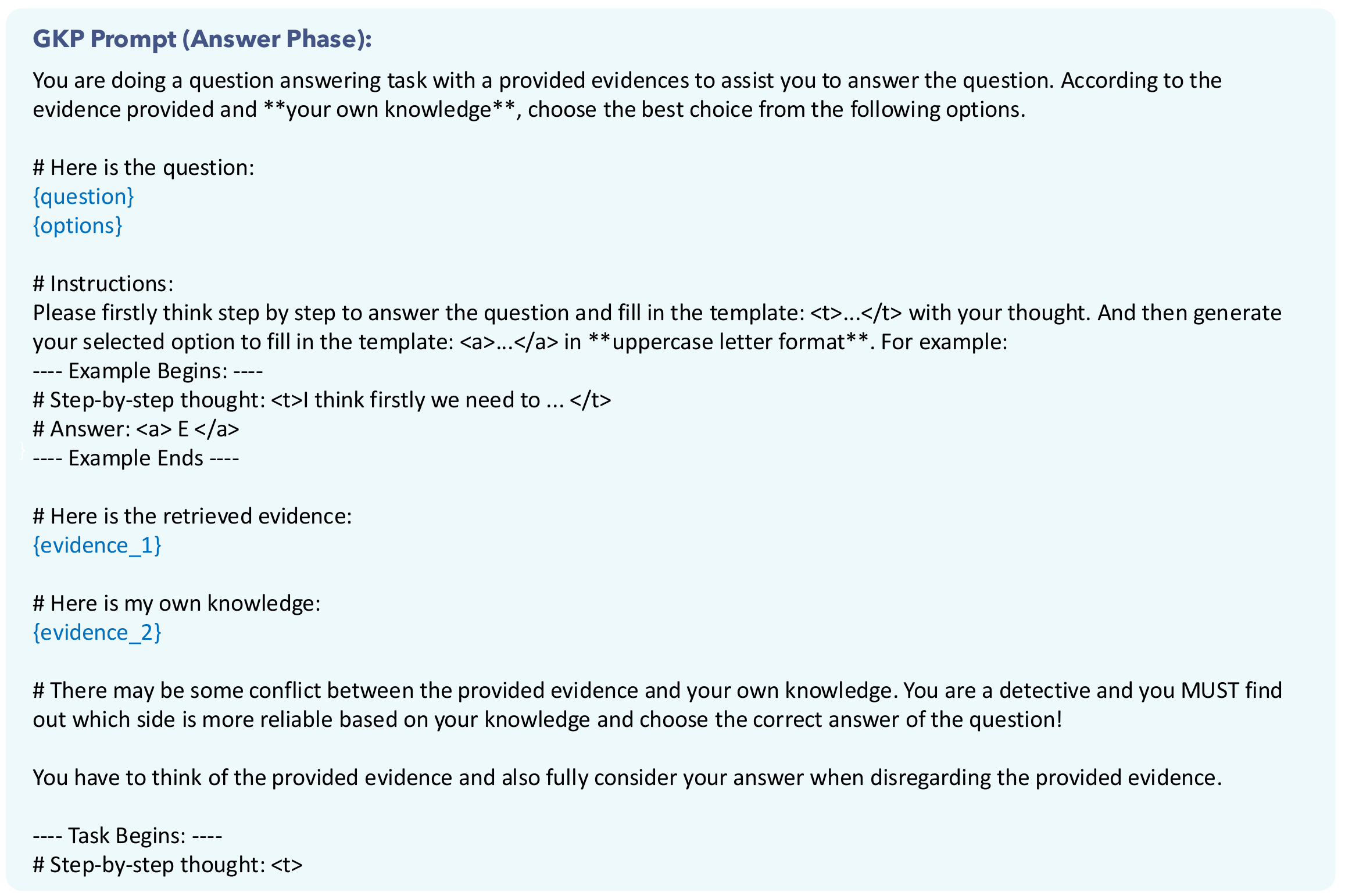}
    \caption{The prompt of the answering phase of GKP~\citep{liu2022generated} baseline method.}
    \label{fig:GKP_ans}
\end{figure*}

\end{document}